\documentclass[10pt]{article}

\PassOptionsToPackage{dvipsnames,table}{xcolor}

\usepackage[preprint]{tmlr_arxiv}

\usepackage{amsmath,amsfonts,bm}

\def\eqref#1{equation~\ref{#1}}

\def\1{\bm{1}}

\DeclareMathAlphabet{\mathsfit}{\encodingdefault}{\sfdefault}{m}{sl}
\SetMathAlphabet{\mathsfit}{bold}{\encodingdefault}{\sfdefault}{bx}{n}

\usepackage{hyperref}
\usepackage{url}
\usepackage{graphicx}
\usepackage{booktabs}
\usepackage{multirow}
\usepackage{makecell}
\usepackage{subcaption}

\usepackage{algorithm}
\usepackage{algpseudocode}

\usepackage{colortbl}
\usepackage{tabularx}
\usepackage{array}
\usepackage{arydshln}   
\usepackage{mathtools}

\usepackage{tikz}
\usepackage{pgfplots}
\usepackage{pgfplotstable}
\pgfplotsset{compat=1.16}
\usepgfplotslibrary{groupplots}
\usetikzlibrary{positioning, shapes, arrows, fit, calc, backgrounds, arrows.meta, shadows}

\usepackage{amsmath,amssymb,amsfonts}
\usepackage{bm}
\usepackage{comment}
\usepackage{soul}
\usepackage{enumitem}
\usepackage{placeins}
\usepackage{float}
\usepackage{pifont}     
\usepackage{caption}

\newcommand{\faCheck}{\ding{51}}
\newcommand{\faTimes}{\ding{55}}
\newcommand{\faBan}{\ding{55}}

\newcommand{\CONE}{\textbf{CM$^{1/2}$}}
\newcommand{\CTWO}{\textbf{CM$^{1/4}$}}
\newcommand{\CTHREE}{\textbf{CM$^{1/9}$}}
\newcommand{\CFOUR}{\textbf{CM$^{1/16}$}}
\newcommand{\CALL}{\textbf{CM$^{all}$}}

\newcommand{\greyrule}{\arrayrulecolor{black!30}\midrule\arrayrulecolor{black}}

\definecolor{pastelbrown}{RGB}{193,154,107}
\definecolor{pastelorange}{RGB}{255,179,102}

\newcommand{\vicc}[1]{{\color{magenta}[vic: #1]}}

\def\month{MM}
\def\year{YYYY}
\def\openreview{\url{https://openreview.net/forum?id=XXXX}}

\title{How Far Can We Go With ImageNet for\\Text-to-Image Generation?}

\author{}

\begin{document}

\maketitle

\vspace{-1.2cm}   
\neuripssetsymbol{equal}{*}
\neuripssetsymbol{equal2}{$\dag$}

\begin{neuripsauthorlist}
\neuripsauthor{Lucas Degeorge}{equal,X,A,I}
\neuripsauthor{Arijit Ghosh}{equal,X,K}
\neuripsauthor{Nicolas Dufour}{X,I,K}
\\
\neuripsauthor{David Picard}{equal2,K}
\neuripsauthor{Vicky Kalogeiton}{equal2,X}
\\
\vspace{0.2cm}
\neuripsauthor{$^1$ \textnormal{LIX, École Polytechnique, CNRS, IP Paris, France}}{}
\neuripsauthor{$^2$ \textnormal{AMIAD}}{}
\neuripsauthor{$^3$ \textnormal{Kyutai}}{}
\neuripsauthor{$^4$ \textnormal{LIGM, École Nationale des Ponts et Chaussées, IP Paris, Univ Gustave Eiffel, CNRS, France}}{}
\end{neuripsauthorlist}

\neuripsaffiliation{X}{LIX, École Polytechnique, CNRS, IP Paris, France}
\neuripsaffiliation{A}{AMIAD}
\neuripsaffiliation{K}{Kyutai}
\neuripsaffiliation{I}{LIGM, École Nationale des Ponts et Chaussées, Univ Gustave Eiffel, CNRS, France}


\vspace{0.4cm}

\begin{abstract}
Recent text-to-image (T2I) generation models have achieved remarkable sucess by training on billion-scale datasets, following a `bigger is better' paradigm that prioritizes data quantity over availability (closed vs open source) and reproducibility (data decay vs established collections). We challenge this established paradigm by demonstrating that one can achieve capabilities of models trained on massive web-scraped collections, using only ImageNet enhanced with well-designed text and image augmentations. With this much simpler setup, we reach the performance of FLUX and achieve a +5 overall score over SD3 on GenEval and +12 on DPGBench over SDXL while using just \emph{ 1/1000th the training images} and $3\times$ to $10\times$ less parameters. This opens the way for more reproducible research as ImageNet is widely available and the proposed standardized training setup only requires 500 hours of H100 to train a text-to-image model.
\end{abstract}

\section{Introduction}
\label{sec:intro}

The prevailing wisdom in text-to-image (T2I) generation holds that larger training datasets inevitably lead to better performance. This ``bigger is better'' paradigm has driven the field to billion-scale image-text paired datasets like LAION-5B~\citep{schuhmann2022laion}, DataComp-12.8B~\citep{gadre2023datacompsearchgenerationmultimodal} or ALIGN-6.6B~\citep{pham2023combinedscalingzeroshottransfer}. 
While this massive scale is often justified as necessary to capture the full text-image distribution, 
in this work, we challenge this assumption and argue that data quantity overlooks fundamental questions of data efficiency and quality in model training.


Our critique of the data-scaling paradigm comes from a critical observation: current training sets are either closed-source or rapidly decaying which makes results impossible to fully reproduce, let alone compare fairly.
As such, the community of T2I generation is in dire need of a standardized training setup to foster open and reproducible research.
Luckily, Computer Vision already has such dataset in ImageNet~\citep{ILSVRC15} that has been the gold standard in many tasks for many years. 
It is widely available and its strength and limitations are well known. 
Moreover, it is heavily used in class-conditional image generation~\citep{peebles2023scalable, jabri2023scalableadaptivecomputationiterative}, which makes its evaluation metrics  more familiar, begging the question of\\
\textit{``\textbf{how far can we go with ImageNet for text-to-image generation?}''}

Our findings are that we can indeed get a surprisingly competitive model by training solely on ImageNet. 
As shown on Figure~\ref{fig:fig_teaser}, we can achieve excellent visual quality. 
Additionally we also achieve very competitive scores on common benchmarks such as GenEval~\citep{ghosh2024geneval} and DPGBench~\citep{hu2024ellaequipdiffusionmodels}, matching or even surpassing popular models that are trained on much more data and at a far greater cost, such as FLUX-dev \cite{labs2025flux1kontextflowmatching}, Stable Diffusion 3 \citep{esser2024scaling}, and Pixart family \citep{chen2023pixart, chen2024pixart} (see Figure~\ref{fig:fig_teaser2}).
However, this does not come without any hurdles. In this paper we analyze the challenges of training using ImageNet only and propose successful strategies to overcome them.
Our strategies allow us to train models from small (about 300M-400M parameters) to larger sizes (1B) on a reasonable compute budget (about 500 H100 hours) making it accessible to more research teams.

Our contributions are thus the following:
\begin{itemize} 
\setlength{\itemsep}{0pt}     
\setlength{\parskip}{0pt}     
\setlength{\parsep}{0pt} 
    \item We analyze the shortcomings of training T2I models on ImageNet and propose mitigation strategies.
    \item Then, we propose a standardized training setup using only images from ImageNet, providing accessible and reproducible research for T2I generation.
    \item We provide several diffusion and flow matching models generating high-quality images and outperforming competitors 10$\times$ larger and trained on 1000$\times$ more data.
\end{itemize}

To commit to open and reproducible science, all our training data will be hosted at <URL> and all our code and models will be hosted at <URL>, publicly online upon acceptance of this paper. 

\input{images/teaser}

\vspace{2mm} 
\noindent \textbf{Positioning of the work.}
We argue that our work is not a replacement for large-scale text-to-image models, but a scientific experiment designed to challenge assumptions about scale. By training only on ImageNet, we move in an orthogonal direction: away from ``bigger is better'' and towards understanding why scale matters. ImageNet provides a well-defined, stable and publicly available corpus that enables controlled, reproducible experimentation, thus setting a common ground for analyzing data efficiency, and generalization in T2I models and promoting open research on this area. 

We posit that scaling is not the only path to progress. Carefully designed experiments, even on limited data, can uncover mechanisms of representation, compositionality, and alignment. We hope our work will inspire the community to look beyond scale and investigate the scientific foundations of T2I generation, such as data efficiency, attribution, and reproducibility, that \emph{complement scale}.

\begin{figure}[ht!]
  \centering
  \begin{tikzpicture}[x=1cm, y=1cm]
    \pgfmathsetmacro{\scaleFactor}{\textwidth / 9cm}
    \begin{scope}[scale=\scaleFactor]
      \node[inner sep=0pt] at (2.25,-1.1) {\begingroup

\pgfplotsset{
    myplotstyle/.style={
    legend style={text=black},
    ylabel style={align=center, font=\tiny},
    xlabel style={align=center, font=\tiny},
    x tick label style={font=\tiny, inner sep=1pt, yshift=10pt, xshift=-3pt},
    y tick label style={font=\tiny, anchor=east, inner sep=1pt},
    x tick style={draw=none},
    scaled ticks=false,
    every axis plot/.append style={ultra thick}, 
    },
    grid = major,
    grid style={dashed, gray!30}
}

\renewcommand\familydefault{\sfdefault}

\begin{tikzpicture}
    \begin{axis}[
        myplotstyle,
        width=8cm, height=5cm,
        ymin=0.4,
        ymax=0.77,
        xlabel={Train Dataset Size (log. scale)},
        ylabel={Geneval Overall scores},
        xmode=log,
        axis background/.style={fill=gray!5},
        ytick={0.4, 0.5, 0.6, 0.7},
        xtick pos=bottom,
    ]

        \addplot[only marks, draw=red, fill=red!40, mark size=3pt, line width=1.25pt] coordinates {
            (1200000,0.61)
        };
        \node[anchor=south west, font=\scriptsize, text=red] at (axis cs:1500000,0.61) {Ours 0.3B};

        \addplot[only marks, draw=red!70!black, fill=red!60, mark size=8pt, line width=1.25pt] coordinates {
            (1200000,0.67)
        };
        \node[anchor=south west, font=\scriptsize, text=red!70!black] at (axis cs:1900000,0.68) {Ours 1B};

        \addplot[only marks, draw=LimeGreen!50!gray, fill=LimeGreen!30, mark size=6pt, line width=1.25pt] coordinates {
            (25500000,0.48)
        } node[anchor=south west, font=\scriptsize, text=LimeGreen] at (axis cs:40000000,0.48) {Pixart-$\alpha$};

        \addplot[only marks, draw=LimeGreen!50!gray, fill=LimeGreen!30, mark size=6pt, line width=1.25pt] coordinates {
            (35500000,0.52)
        } node[anchor=south west, font=\scriptsize, text=LimeGreen] at (axis cs:52000000,0.52) {Pixart-$\Sigma$};

        \addplot[only marks, draw=Peach!50!gray, fill=Lavender!10, mark size=3pt, line width=1.25pt] coordinates {
            (20000000,0.75)
        } node[anchor=east, font=\scriptsize, text=Lavender] at (axis cs:98000000,0.75) {MIRO};

        \addplot[only marks, draw=teal, fill=teal!40, mark size=10pt, clip mode=individual, line width=1.25pt] coordinates {
            (5850000000,0.55)
        } node[anchor=south east, font=\scriptsize, text=teal] at (axis cs:3000000000,0.55) {SDXL};

        \addplot[only marks, draw=violet, fill=violet!40, mark size=7pt, clip mode=individual, line width=1.25pt] coordinates {
            (5850000000,0.5)
        } node[anchor=south east, font=\scriptsize, text=violet] at (axis cs:3500000000,0.5) {SD2.1};

        \addplot[only marks, draw=orange!70!black, fill=orange!40, mark size=8pt, clip mode=individual, line width=1.25pt] coordinates {
            (1000000000,0.62)
        } node[anchor=south east, font=\scriptsize, text=orange!70!black] at (axis cs:600000000,0.63) {SD3 M};

        \addplot[only marks, draw=Cyan!60!black, fill=Cyan!30, mark size=6pt, line width=1.25pt] coordinates {
            (5850000000,0.64)
        } node[anchor=south east, font=\scriptsize, text=Cyan!60!black] at (axis cs:18000000000,0.60) {SANA-0.6};

        \addplot[only marks, draw=gray!70!black, fill=gray!40, mark size=14pt, clip mode=individual, line width=1.25pt] coordinates {
            (10000000000,0.67)
        } node[anchor=south west, font=\scriptsize, text=gray!70!black] at (axis cs:500000000,0.685) {FLUX-dev};

        \addplot[only marks, draw=red, fill=red, mark size=0.25pt] coordinates {
           (1200000,0.61)
        };

        \addplot[only marks, draw=red!70!black, fill=red!70!black, mark size=0.25pt] coordinates {
           (1200000,0.67)
        };

        \addplot[only marks, draw=LimeGreen, fill=LimeGreen, mark size=0.25pt] coordinates {
            (25500000,0.48)
        };
        \addplot[only marks, draw=LimeGreen, fill=LimeGreen, mark size=0.25pt] coordinates {
            (35500000,0.52)
        };

        \addplot[only marks, draw=Peach, fill=Lavender, mark size=0.25pt] coordinates {
            (20000000,0.75)
        };

        \addplot[only marks, draw=teal, fill=teal, mark size=0.25pt] coordinates {
            (5850000000,0.55)
        };

        \addplot[only marks, draw=violet, fill=violet, mark size=0.25pt] coordinates {
            (5850000000,0.5)
        };

        \addplot[only marks, draw=orange!70!black, fill=orange!70!black, mark size=0.25pt] coordinates {
            (1000000000,0.62)
        };

        \addplot[only marks, draw=Cyan!60!black, fill=Cyan!60!black, mark size=0.25pt] coordinates {
            (5850000000,0.64)
        };

        \addplot[only marks, draw=gray!70!black, fill=gray!70!black, mark size=0.25pt] coordinates {
            (10000000000,0.67)
        };
    \end{axis}
\end{tikzpicture}

\endgroup};
      \node[inner sep=0pt] at (6.75,-1.1) {\begingroup

\pgfplotsset{
    myplotstyle/.style={
    legend style={text=black},
    ylabel style={align=center, font=\tiny},
    xlabel style={align=center, font=\tiny},
    x tick label style={font=\tiny, inner sep=1pt, yshift=10pt, xshift=-3pt},
    y tick label style={font=\tiny, anchor=east, inner sep=1pt},,
    x tick style={draw=none},
    scaled ticks=false,
    every axis plot/.append style={ultra thick}, 
    },
    grid = major,
    grid style={dashed, gray!30}
}

\renewcommand\familydefault{\sfdefault}

\begin{tikzpicture}
    \begin{axis}[
        myplotstyle,
        width=8cm, height=5cm,
        ymin=0.55,
        ymax=0.89,
        xlabel={Train Dataset Size (log. scale)},
        ylabel={DPG Benchmark Overall scores},
        xmode=log,
        axis background/.style={fill=gray!5},
        ytick={0.6, 0.7, 0.8},
        xtick pos=bottom
    ]

        \addplot[only marks, draw=red!70!black, fill=red!60, mark size=7pt, line width=1.25pt] coordinates {
            (1200000,0.8117)
        };
        \node[anchor=south west, font=\scriptsize, text=red!70!black] at (axis cs:1900000,0.815) {Ours 1B};

        \addplot[only marks, draw=LimeGreen!50!gray, fill=LimeGreen!30, mark size=6pt, line width=1.25pt] coordinates {
            (25500000,0.71)
        } node[anchor=south west, font=\scriptsize, text=LimeGreen] at (axis cs:40000000,0.71) {Pixart-$\alpha$};

        \addplot[only marks, draw=LimeGreen!50!gray, fill=LimeGreen!30, mark size=6pt, line width=1.25pt] coordinates {
            (35500000,0.795)
        } node[anchor=south west, font=\scriptsize, text=LimeGreen] at (axis cs:52000000,0.795) {Pixart-$\Sigma$};

        \addplot[only marks, draw=Peach!50!gray, fill=Peach!30, mark size=5pt, line width=0.5pt] coordinates {
             (20000000,0.50)
        } node[anchor=south west, font=\scriptsize, text=Peach] at (axis cs:20000000,0.5) {CAD};

        \addplot[only marks, draw=teal, fill=teal!40, mark size=10pt, clip mode=individual, line width=1.25pt] coordinates {
            (5850000000,0.75)
        } node[anchor=south east, font=\scriptsize, text=teal] at (axis cs:3000000000,0.73) {SDXL};

        \addplot[only marks, draw=violet, fill=violet!40, mark size=7pt, clip mode=individual, line width=1.25pt] coordinates {
            (5850000000,0.63)
        } node[anchor=south east, font=\scriptsize, text=violet] at (axis cs:3500000000,0.63) {SD 1.5};

        \addplot[only marks, draw=orange!70!black, fill=orange!40, mark size=8pt, clip mode=individual, line width=1.25pt] coordinates {
            (1000000000,0.841)
        } node[anchor=south east, font=\scriptsize, text=orange!70!black] at (axis cs:600000000,0.84) {SD3 M};

        \addplot[only marks, draw=Cyan!60!black, fill=Cyan!30, mark size=6pt, line width=1.25pt] coordinates {
            (5850000000,0.843)
        } node[anchor=north, font=\tiny, text=Cyan!70!black] at (axis cs:2600000000,0.83) {SANA};

        \addplot[only marks, draw=gray!70!black, fill=gray!40, mark size=14pt, clip mode=individual, line width=1.25pt] coordinates {
            (10000000000,0.84)
        } node[anchor=south east, font=\tiny, text=gray!70!black] at (axis cs:5000000000,0.875) {FLUX-dev};

        \addplot[only marks, draw=red!70!black, fill=red!70!black, mark size=0.25pt] coordinates {
            (1200000,0.8117)
        };

        \addplot[only marks, draw=LimeGreen, fill=LimeGreen, mark size=0.25pt] coordinates {
            (25500000,0.71)
        };
        \addplot[only marks, draw=LimeGreen, fill=LimeGreen, mark size=0.25pt] coordinates {
            (35500000,0.795)
        };

        \addplot[only marks, draw=teal, fill=teal, mark size=0.25pt] coordinates {
            (5850000000,0.75)
        };

        \addplot[only marks, draw=violet, fill=violet, mark size=0.25pt] coordinates {
            (5850000000,0.63)
        };

        \addplot[only marks, draw=orange!70!black, fill=orange!70!black, mark size=0.25pt] coordinates {
            (1000000000,0.841)
        };

        \addplot[only marks, draw=Cyan!60!black, fill=Cyan!60!black, mark size=0.25pt] coordinates {
            (5850000000,0.843)
        };

        \addplot[only marks, draw=gray!70!black, fill=gray!70!black, mark size=0.25pt] coordinates {
            (10000000000,0.84)
        };

    \end{axis}
\end{tikzpicture}

\endgroup};
    \end{scope}
  \end{tikzpicture}
  \caption{\textbf{Quantitative results} on GenEval (left) and DPGBench (right).
    The size of the bubble represents the number of parameters.
    In both cases, we outperform models of $10\times$ the parameters trained
    on $1000\times$ the number of images.}
  \label{fig:fig_teaser2}
\end{figure}


\section{Related Work}
\label{sec:related}


\subsection{Background}
\label{subsec:background}

\paragraph{Diffusion Models.}~\cite{song2020score, ho2020denoising, sohl2015deep} have demonstrated remarkable success across various domains~\cite{huang2023noise2music,courant2025exceptional, dufour2024around}. While image generation remains their most prominent application~\cite{dhariwal2021diffusion, song2020score, karras2022elucidating}, text-to-image (T2I) synthesis~\cite{rombach2022high, saharia2022photorealistic, ramesh2022hierarchical} has emerged as a particularly impactful use case. These models operate by learning to reverse a gradual Gaussian noise corruption process. At extreme noise levels, the model effectively samples from a standard normal distribution to produce realistic images. The core optimization objective is:
\\
\begin{equation}
    \min_{\theta} \mathbb{E}_{(x_0,c) \sim p_\text{data}, \epsilon \sim \mathcal{N}(0,1)} \left[ \left\| \epsilon - \epsilon_{\theta}(x_t, c, t) \right\|^2 \right]
    \label{eq:diff}
\end{equation}
\\
where $x_t {=} \sqrt{\gamma(t)} x_0 {+} \sqrt{1-\gamma(t)} \epsilon$ denotes the noised image at timestep $t$, $x_0$ the original image, $c$ the corresponding condition (such as text), $\epsilon$ is standard normal noise, $\epsilon_{\theta}$ the learned noise predictor, and $\gamma(t)$ the variance schedule.

\vspace{2mm} \noindent \textbf{Flow Matching.} Flow matching~\citep{lipman2022flow, liu2022rectifiedflow} offers an alternative to the diffusion formulation above, framing generation as learning a continuous-time velocity field along a path between the data distribution and a Gaussian prior, rather than reversing a fixed noising process. Using the same notation as Equation~\ref{eq:diff}, a common choice is the linear interpolant $x_t {=} (1{-}t) x_0 {+} t \epsilon$, whose constant velocity $\epsilon {-} x_0$ is regressed by the model:
\\
\begin{equation}
    \min_{\theta} \mathbb{E}_{(x_0,c) \sim p_\text{data}, \epsilon \sim \mathcal{N}(0,1), t} \left[ \left\| (\epsilon - x_0) - v_{\theta}(x_t, c, t) \right\|^2 \right]
    \label{eq:fm}
\end{equation}
\\
where $v_{\theta}$ is the learned velocity field. Sampling then integrates this velocity field with an ODE solver, e.g.\ Euler~\citep{ma2024eccvSIT} or Heun~\citep{li2025jit}.

\vspace{2mm} \noindent \textbf{Architectures.} Recent T2I models increasingly replace the convolutional U-Net backbone~\citep{ho2020denoising, rombach2022high} with plain transformers operating on latent or pixel patches. DiT~\citep{peebles2023scalable} showed that such a transformer, trained with the diffusion objective (Equation~\ref{eq:diff}), scales favorably with compute and has since been adopted by several T2I systems~\citep{chen2023pixart, esser2024scaling, chen2024pixart}. SiT~\citep{ma2024eccvSIT} keeps the same backbone but trains it with the flow-matching objective (Equation~\ref{eq:fm}) instead, a recipe later scaled to billions of parameters for T2I by Stable Diffusion 3~\citep{esser2024scaling}. RIN~\citep{jabri2023scalableadaptivecomputationiterative} iteratively routes computation between a small set of latent tokens and the full set of image tokens to reduce the cost of processing long sequences; \citet{dufour2024don} adapts RIN for text-conditional generation into the TextRIN architecture. Rather than compressing images into a latent space~\citep{rombach2022high}, a growing line of work denoises raw pixels directly: JiT~\citep{li2025jit} patchifies pixels for a plain transformer, PixNerd~\citep{wang2025pixnerd} decodes each patch with a neural field, PixelDiT~\citep{yu2026pixeldit} splits the backbone into a patch-level and a pixel-level DiT, and PixelGen~\citep{ma2026pixelgen} adds a perceptual loss.

\subsection{Related Work}
\label{subsec:related_work}

\vspace{2mm} \noindent \textbf{Computational Efficiency.} Early diffusion models consumed hundreds of thousands of GPU hours~\cite{dhariwal2021diffusion, nichol2021improved}. Training efficiency improved drastically by operating in compressed latent spaces~\cite{rombach2022high}, which enabled subsequent architectural advances~\cite{peebles2023scalable,jabri2023scalableadaptivecomputationiterative}. Building on this, \cite{xie2024sana} introduced an even more compressed latent space to further streamline text-to-image models. Addressing the limitations of standard diffusion loss, \cite{wei2023diffusion, yu2024representation} demonstrated that supplementary representation losses accelerate convergence. Similarly, \cite{zheng2025diffusion} leveraged a self-supervised autoencoder to achieve state-of-the-art class-conditional generation at a fraction of the usual cost. Pushing this concept further, \cite{zheng2025rae} replaced the standard VAE with a representation autoencoder---combining a frozen vision encoder (DINO, SigLIP, or MAE) with a trained decoder---proving that a semantically structured, high-dimensional latent space provides a stronger substrate for diffusion transformers. Compute has also been optimized via token routing, which dynamically allocates budget to specific tokens to improve both convergence and efficiency~\cite{krause2025tread, park2025sprint}. \cite{chen2023pixart} achieved dramatic savings by repurposing class-conditional models for text-to-image tasks, a method later enhanced with higher-quality data~\cite{chen2024pixart}. Additionally, \cite{dufour2024don} utilized coherence-aware mechanisms to match Stable Diffusion's performance~\cite{rombach2022high} using 100$\times$ fewer GPU hours. Beyond training, extensive parallel efforts have reduced inference costs~\cite{frans2024one, geng2025mean, wimbauer2024cache, ma2024learning, ma2024deepcache}.

\vspace{2mm} \noindent \textbf{Data Efficiency.} T2I models typically rely on billion-scale web-scraped datasets~\citep{rombach2022high}, creating storage and copyright barriers. Pioneering data curation, \cite{chen2023pixart} utilized 20M high-quality images from recaptioned SAM data~\citep{kirillov2023segment}, though portions remain proprietary. \cite{shen2025mmdit} later demonstrated that strong models can be trained on a 25M public multi-source dataset.  Works using CC12M~\citep{changpinyo2021cc12m, gu2023matryoshka, dufour2024don} and YFCC100M's public subset~\citep{thomee2016yfcc100m, gokaslan2024commoncanvas} have successfully trained models on 10M and 100M public images. However, the regime below 10M images has never been explored.

Pushing toward full reproducibility, recent works train competitive models exclusively on public data. \cite{wang2026minit2i} trained a minimal 258M-parameter pixel-space flow-matching model. Scaling up, \cite{zeng2026i1} released a 3B model distilled from over 300 controlled experiments, and \cite{aubin2026monet} curated MONET, an openly licensed 105M image-text collection filtered and recaptioned from 2.9B raw pairs. Other works question the necessity of raw data volume: \cite{ding2025alchemist} used a meta-gradient criterion to match full-dataset performance with half the data, and \cite{chen2026lens} traded volume for caption density using 800M heavily recaptioned pairs.

Our approach diverges in its data requirements. Instead of curating or subsampling massive corpora, we transform the standard and reproducible ImageNet~\citep{ILSVRC15} dataset into T2I training data via image and text augmentations. Notably, the design choices of \cite{chen2026lens} and \cite{wang2026minit2i} remain complementary to ours and could be directly applied to our setting.

\vspace{2mm} \noindent \textbf{Joint Efficiency.} A third research line examines the interaction between data and compute. \cite{liang2024scaling} derives scaling laws for diffusion transformers, linking compute, model, and dataset sizes to predict optimal resource allocation. Extending this to T2I flow-matching transformers, \cite{chickering2026abra} finds that compute optimality requires roughly $10\times$ more data per parameter than language models. Importantly for our setting, they show these models are highly robust to overtraining, making repeated passes over small corpora viable. Alternatively, MIRO~\citep{dufour2025miro} addresses this trade-off by conditioning pretraining on multiple reward signals to achieve target quality with significantly less compute. Complementing these works, we investigate how far a single, small dataset (ImageNet) can push T2I quality at a fixed model size and compute budget using augmentations, rather than scaling data.

\section{Experimental Setup}
\label{sec:setup}

\subsection{Architectures}
\label{subsec:architectures}

We experiment with transformer-based generative architectures spanning three modeling families: latent-space diffusion, latent-space flow matching, and pixel-space flow matching.

\vspace{2mm} \noindent \textbf{Latent-space diffusion.} Both models below are trained in the latent space of the Stable Diffusion VAE~\citep{rombach2022high} with the standard diffusion objective (Equation~\ref{eq:diff}).
\begin{itemize}
\setlength{\itemsep}{2pt}
    \item \textbf{DiT-I}: our adaptation of DiT~\citep{peebles2023scalable} for text-conditional generation. We replace \texttt{AdaLN-Zero} conditioning with \textit{cross-attention} to inject text, as in PixArt-$\alpha$~\citep{chen2023pixart}, and refine the text condition with two self-attention layers before cross-attention. Following~\citet{esser2024scaling}, we add QK-Normalization~\citep{henry2020query} to all self-attention and cross-attention blocks to reduce training instability, and LayerScale~\citep{touvron2021going} to all residual blocks.
    \item \textbf{TextRIN-I}: the off-the-shelf TextRIN architecture from~\citet{dufour2024don}, an adaptation of RIN~\citep{jabri2023scalableadaptivecomputationiterative} for text-conditional generation.
\end{itemize}

\vspace{2mm} \noindent \textbf{Latent-space flow matching.} \textbf{SiT-I} uses the same backbone and text-conditioning adaptation as DiT-I (cross-attention injection, QK-Normalization, LayerScale), but is trained with a flow-matching objective~\citep{ma2024eccvSIT} instead of diffusion, using log-normal timestep sampling and an Euler ODE sampler at inference.

\vspace{2mm} \noindent \textbf{Pixel-space flow matching.} \textbf{JiT-I} operates directly on raw pixels, without a VAE. We adapt JiT~\citep{li2025jit} for text-conditional generation following the recipe of PixelGen~\citep{ma2026pixelgen}
The model is trained with a flow-matching objective and sampled with a Heun ODE solver.

The suffix ``I'' denotes that the models are trained only on ImageNet. More details about the architectures and the training recipe are provided in Appendix~\ref{app:train_arch_details}.

\subsection{Evaluation Metrics}
\label{subsec:metrics}

We evaluate our models along three axes:

\vspace{2mm} \noindent \textbf{Image quality.} We assess generation quality by comparing generated images to both in-domain (ImageNet-50k validation set) and out-of-domain (MS-COCO-30k~\citep{lin2014microsoft}) reference distributions. We adopt: (1) FID~\citep{heusel2017gans} with both \texttt{Inception-v3}~\citep{szegedy2016rethinking} and \texttt{DINOv2}~\citep{oquab2023dinov2} backbones; (2) Precision, (3) Recall~\citep{kynkaanniemi2019improved}; (4) Density, (5) Coverage~\citep{naeem2020reliable}. The latter four use \texttt{DINOv2} features.

\vspace{2mm} \noindent \textbf{Compositionality.} To assess text-image alignment and compositional generation, we use: (1) CLIPScore~\citep{hessel2021clipscore} and (2) Jina-CLIP Score~\citep{koukounas2024jina} on both MS-COCO-30k and ImageNet validation sets; (3) GenEval~\citep{ghosh2024geneval}; (4) DPGBench~\citep{hu2024ellaequipdiffusionmodels}.

\vspace{2mm} \noindent \textbf{Aesthetic quality.} To assess human aesthetic alignment, following~\citet{eyring2024reno}, we adopt: (1) PickScore~\citep{Kirstain2023PickaPicAO}; (2) Aesthetics Score~\citep{schuhmann2022laion}; (3) HPSv2.1~\citep{wu2023human}; (4) ImageReward~\citep{xu2023imagereward}, all measured on the \texttt{PartiPrompts}~\citep{yu2022scalingautoregressivemodelscontentrich} test bench.



\section{Adopting ImageNet for Text-to-Image Generation}
\label{sec:method}

We focus on training T2I models using ImageNet, a small, open-source and widely accepted data collection. We first discuss the evaluation criterions and then gradually pinpoint the major limitations in setting up a T2I diffusion model using ImageNet. To overcome these limitations, we show that well-crafted augmentations can bring forth a compositionally accurate T2I model while training under hard data constraints.

\subsection{Text challenges}

\paragraph{Challenge: \textit{Absence of captions.}} 
Class-conditional models trained with ImageNet have shown exceptional generation capabilities~\citep{peebles2023scalable,ma2024eccvSIT}. However, extending this use to T2I generation is difficult since ImageNet, being a classification dataset, lacks any sort of caption corresponding to its images. To adopt ImageNet for T2I generation, similar to prior works~\citep{radford2021learning}, one could build captions by a very simple strategy of \texttt{`An image of <class name>'} (denoted AIO). However, this results in very poor generation capabilities as shown in Table \ref{tab:1_aio_TA}. 

This low generalization can be attributed to the three major shortcomings of AIO captions for ImageNet: 
First, AIO captions lack vocabulary. They contain only roughly a thousand words corresponding to the concepts of the classes and thus lack attributes, spatial relations, etc. This constraint on the diversity in the text-condition space leads to a clear bottleneck in text understanding. 
Second, there is often more content in the image than just the class. For example, a caption ``an image of \textit{golden retriever}'' mentions the class name but leaves out details and concepts that could be in the background. This lack of details leads to spurious correlation where the model can learn to associate unrelated visual pattern (e.g., grass texture) with the class name (e.g., \textit{golden retriever}) because the text for this concept is never mentioned in the text space. 
Third, despite the presence of humans in the images, ImageNet does not contain a `\textit{person}' class, resulting in humans not being represented in the AIO text space. This issue extends to many categories (\emph{road}, \emph{water}, etc), as ImageNet is an object-centric dataset.

\vspace{2mm}
\noindent \textbf{Solution: Long informative captions (TA).}
To overcome this, we employ a synthetic captioner~\citep{liu2024visual} (\emph{Text Augmentation}, TA) to generate comprehensive captions that capture: (i)~scene composition and spatial relationships; (ii)~background elements and environmental context; (iii)~secondary objects and participants; (iv)~visual attributes (color, size, texture); and (v)~actions and interactions. More details about the captioning are given in Appendix~\ref{app:captions}


\begin{table}
    \centering
    \resizebox{\linewidth}{!}{
        \begin{tabular}{l@{\;\;}c!{\color{gray!60}\vrule}c@{\;}c@{\;}c@{\;}c@{\;}c@{\;}c@{\;}!{\color{gray!60}\vrule}c@{\;}c@{\;}c@{\;}}
            \toprule
            \multirow{2}{*}{\textbf{Model}} & \multirow{2}{*}{\textbf{TA}} &
            \multicolumn{2}{c}{\textbf{FID Inc.}$\downarrow$} &  \multirow{2}{*}{\colorbox{YellowOrange!20} {\textbf{Prec.}$\uparrow$}} & \multirow{2}{*}{\colorbox{YellowOrange!20} {\textbf{Rec.}$\uparrow$}} & \multirow{2}{*}{\colorbox{YellowOrange!20} {\textbf{Den.}$\uparrow$}} & \multirow{2}{*}{\colorbox{YellowOrange!20} {\textbf{Cov.}$\uparrow$}} & \multicolumn{2}{c}{\textbf{Jina CLIP}$\uparrow$} &\multirow{2}{*}{\textbf{GenEval}$\uparrow$}  \\
            & & \colorbox{YellowOrange!20} {\small \textbf{IN-Val}} & \colorbox{Orchid!20} {\small \textbf{COCO}} & & & & & \colorbox{YellowOrange!20} {\small \textbf{IN-Val}} & \colorbox{Orchid!20} {\small \textbf{COCO}} & \\
            \midrule
             & \textcolor{Red}{\faTimes} & 20.14 & 71.00 & 0.67  & 0.29   & 0.71  & 0.39 & 31.21 & 22.42 & 0.11 \\
            \multirow{-2}{*}{DiT-I} & \textcolor{Green}{\faCheck} & 6.29 & 45.71 & 0.77 & 0.76 & 0.82 & 0.72 & 38.45 & 38.39 & 0.55 \\ 
            \greyrule
            & \textcolor{Red}{\faTimes} & 84.77 & 46.35 &  0.75 & 0.05 & 1.40 & 0.10 & 20.55 & 14.06 & 0.17  \\
            \multirow{-2}{*}{TextRIN-I} & \textcolor{Green}{\faCheck} & 6.16 & 49.89 &  0.80 & 0.72 & 0.89 & 0.76 & 38.01 & 37.85 & 0.55 \\
            \bottomrule\\
        \end{tabular}
        }
    \caption{\textbf{Image quality and compostionality} of AIO models (\textcolor{Red}{\faTimes}) and TA models (\textcolor{Green}{\faCheck}). FID reported is FID \texttt{Inception v3}. Precision, Recall, Density and Coverage are computed using \texttt{DINOv2} features on \colorbox{YellowOrange!20}{ImageNet Val}. Results on \colorbox{Orchid!20}{COCO test} set in suppl.} 
    \label{tab:1_aio_TA}
\end{table}

We compare the gains attributed to long captions quantitatively in Table \ref{tab:1_aio_TA}. For ImageNet-Val set, we observe that models trained with long captions significantly improves performance, resulting in lower FIDs of \textbf{6.29} for DiT-I and \textbf{6.16} for TextRIN-I in contrast to \textbf{20.14} for DiT-I and \textbf{85} for TextRIN-I on AIO captions. As a point of reference, we remind the reader that models of this size (below 0.5B parameters) typically have an FID of $9$ using the class-conditional setup~\citep{peebles2023scalable}. 
Additional evidence of image quality improvement is found with the Precision (P), Recall (R), Density (D), and Coverage (C) metrics. DiT-I achieves better performance with text-augmentation over AIO on all four of the P,R,D,C metrics, whereas, CAD-I shows improvement in three out of four of these metrics, indicating that TA is quite vital for transforming ImageNet into a T2I specific dataset. 
For COCO test set which is a zero-shot task for our training, this trend is all the more dramatic. The TA models are the only ones able to correctly follow the prompt as attested by the much improved Jina CLIP score (DiT-I from \textbf{22.42} to \textbf{38.45}; TextRIN-I from \textbf{14.06} to \textbf{37.85}), while keeping similar image quality. 

Regarding text-image alignment and compositionality, models trained with longer captions benefit from the added information, evidenced by the improvement in GenEval overall score from (DiT-I from \textbf{0.11} to \textbf{0.55}; TextRIN-I from \textbf{0.17} to \textbf{0.55}) (see Table \ref{tab:1_aio_TA}, last column).

\subsection{Image challenges}
\label{subsec:images_aug}

Using long, informative captions (TA) significantly enhances both generation quality and compositional alignment. But training text-to-image diffusion models on ImageNet only still faces two critical limitations: early overfitting and poor compositional generalization.

\vspace{2mm} \noindent \textbf{Limitation 1: \textit{Early overfitting}.}
Models trained on ImageNet with long captions (TA) demonstrate promising initial performance. However, due to the relatively small scale of ImageNet (only 1.2 million images) they begin overfitting at approximately 200k training steps (see Figure \ref{fig:overfitting}). Interestingly, this overfitting is specific to latent-space architectures. On the contrary, the pixel-space JiT-I does not degrade over the same training budget. We attribute this to the VAE acting as a perceptual bottleneck: by discarding high-frequency and pixel-level information, it lowers the effective entropy of the denoising target, making it easier for a fixed-capacity model to memorize a small dataset. JiT-I is not afforded this simplification.

\vspace{2mm} \noindent \textbf{Limitation 2: \textit{Restricted Complex Compositionality Abilities}.}
ImageNet's object-centric nature presents a challenge for learning complex compositions. Even with enhanced textual descriptions via TA captioning, models still struggle with spatial relationships, attribute binding, and multi-object compositions. This limitation manifests in lower GenEval scores for compositional prompts involving multiple objects, color attribution, and positional relationships, as shown in Table \ref{tab:geneval_plus_aesthetic}.



\begin{figure}[t]
    \centering
    \resizebox{\linewidth}{!}{
    \pgfplotsset{
    mylegendbox/.style={
        text=black, font=\fontsize{4}{4.5}\selectfont,
        fill=white, fill opacity=1, draw=gray!40, rounded corners=1pt,
        inner sep=1.5pt, row sep=-1pt,
    },
    myplotstyle/.style={
        legend style={mylegendbox, at={(0.98,0.98)}, anchor=north east},
        legend cell align={left},
        legend image post style={scale=0.4},
        ylabel style={align=center, font=\tiny},
        xlabel style={align=center, font=\tiny},
        title style={font=\small},
        x tick label style={font=\tiny, inner sep=1pt, yshift=6pt},
        y tick label style={font=\tiny, inner sep=1pt},
        xtick align=inside,
        scaled ticks=false,
        every axis plot/.append style={thick, mark=*, mark size=1pt},
    },
    grid = major,
    grid style={dashed, gray!30}
}

\begin{tikzpicture}
    \begin{groupplot}[
        group style={
            group size=3 by 1,
            horizontal sep=1cm
        },
        width=5cm,
        height=4.2cm,
        axis lines=box,
        axis background/.style={fill=gray!5}
    ]

    \nextgroupplot[
        myplotstyle,
        title={TextRIN-I},
        xlabel={\tiny Training Steps (k)},
        ylabel={\tiny GenEval score},
        ylabel near ticks,
    ]
    \addplot[CarnationPink!80] coordinates {
        (100,0.53) (150,0.55) (200,0.54) (250,0.56) (300,0.54)
        (350,0.52) (400,0.50) (450,0.49) (500, 0.484)
    };
    \addlegendentry{TA}

    \addplot[JungleGreen!80] coordinates {
        (100,0.51) (150,0.548) (200,0.552) (250,0.553) (300,0.548)
        (350,0.53) (400,0.53) (450,0.52) (500, 0.53)
    };
    \addlegendentry{TA+IA}

    \nextgroupplot[
        myplotstyle,
        title={SiT-I},
        xlabel={\tiny Training Steps (k)},
        ylabel={\tiny GenEval score},
        ylabel near ticks,
        xmax=305,
        xtick={0,150},
        legend style={mylegendbox, at={(0.98,0.5)}, anchor=east},
    ]
    \addplot[CarnationPink!80] coordinates {
        (25,0.43472) (50,0.48169) (75,0.51241) (100,0.4957) (125,0.49112)
        (150,0.47687) (175,0.45878) (200,0.44417) (250,0.43094) (275,0.40352)
        (300,0.41376) (325,0.40215) (350,0.38598) (375,0.3809) (400,0.37078)
    };
    \addlegendentry{TA}

    \addplot[JungleGreen!80] coordinates {
        (25,0.43492) (50,0.53621) (75,0.56312) (100,0.58727) (125,0.61006)
        (150,0.60185) (175,0.59183) (200,0.59557) (225,0.59359) (250,0.59484)
        (275,0.59592) (300,0.58606) (325,0.59272)
    };
    \addlegendentry{TA+IA}

    \nextgroupplot[
        myplotstyle,
        title={JiT-I},
        xlabel={\tiny Training Steps (k)},
        ylabel={\tiny GenEval score},
        ylabel near ticks,
        xmax=305,
        xtick={0,150},
        legend style={mylegendbox, at={(0.98,0.02)}, anchor=south east},
    ]
    \addplot[CarnationPink!80] coordinates {
        (25,0.29422) (50,0.42692) (75,0.48516) (100,0.50817) (125,0.51483)
        (150,0.5289) (175,0.53202) (200,0.53092) (225,0.5357) (250,0.5315)
        (275,0.52562) (300,0.52373) (325,0.52215) (350,0.51883) (375,0.51148) (400,0.52738)
    };
    \addlegendentry{TA}

    \addplot[JungleGreen!80] coordinates {
        (25,0.22768) (50,0.33768) (75,0.41599) (100,0.43947) (125,0.45835)
        (150,0.47825) (175,0.48596) (200,0.49105) (225,0.49236) (250,0.50506)
        (275,0.50984) (300,0.50857)
    };
    \addlegendentry{TA+IA}

    \end{groupplot}
\end{tikzpicture}}
    \caption{\textbf{Training dynamics showing GenEval scores vs training steps}, for TextRIN-I~\cite{dufour2024don}, SiT-I~\cite{ma2024eccvSIT}, and JiT-I~\cite{li2025jit}. \colorbox{JungleGreen!40}{\textbf{TA + IA}} maintains better scores throughout training compared to \colorbox{CarnationPink!40}{\textbf{TA}} only and improves resistance to overfitting for the latent architectures.}
    \label{fig:overfitting}
\end{figure}

\vspace{2mm} \noindent \textbf{Solution: \textit{Image Augmentation (IA)}.} 
To reduce overfitting and improve compositional reasoning, we investigate the use of image augmentations during training. We experiment with two augmentation strategies. Details about implementation and training pipeline are given in supl.\ material.

\begin{itemize}
    \item\colorbox{Cerulean!20}{\textbf{CutMix}}~\citep{yun2019cutmix}: For each image, we randomly select an image from a different class and overlay a smaller version of it onto the original image. A caption is generated using the CutMix image as input. This technique introduces additional variability in the training data. 
    \item\colorbox{Lavender!20}{\textbf{Crop}}: During training, we randomly mask a portion of the image tokens such that the model is exposed only to a local crop of the original image. We add crop coordinates tokens to the captions of the image. It encourages the model to decouple object features from their background context, and to learn correspondences between partial visual elements and specific text tokens. 
\end{itemize}

\paragraph{Modified training loss for CutMix.}
Because CutMix images have sharp composition boundaries, training on them at all noise levels would cause the model to reproduce the boundary artifacts. We prevent this with a threshold-gated conditional sampling scheme that only exposes CutMix images once the corrupted input is noisy enough that the artifact is no longer visible, applied to both the diffusion loss (DiT-I, TextRIN-I) and the flow-matching loss (SiT-I). Full equations for both objectives, and the corresponding batch-construction algorithm, are given in Appendix~\ref{app:cutmix}.

\begin{table*}[]
\centering
\resizebox{\linewidth}{!}{
\begin{tabular}{l c @{\;}!{\color{gray!60}\vrule} c c c c c c c @{\;}!{\color{gray!60}\vrule} c c c c}
    \toprule

    \multirow{2}{*}{\textbf{Model}} &
    \multirow{2}{*}{\textbf{IA}} &
    \multicolumn{7}{c!{\color{gray!60}\vrule}}{\textbf{GenEval}} &
    \multicolumn{4}{c}{\textbf{Aesthetic metrics}} \\

    \cmidrule(lr){3-9} \cmidrule(lr){10-13}

    &
    & \textbf{Overall$\uparrow$} & \textbf{One obj.$\uparrow$} & \textbf{Two obj.$\uparrow$} & \textbf{Count.$\uparrow$} & \textbf{Col.$\uparrow$} & \textbf{Pos.$\uparrow$} & \textbf{Col. attr.$\uparrow$}
    & \textbf{PickScore$\uparrow$} & \textbf{Aes.Score$\uparrow$} & \textbf{HPSv2.1$\uparrow$} & \textbf{ImageReward$\uparrow$} \\

    \midrule

    {} & \textcolor{red}{\faTimes}
        & 0.55 & 0.95 & 0.61 & 0.36 & 0.80 & 0.28 & 0.33
        & 20.74 & 5.29 & 0.25 & 0.18 \\

    {} & \colorbox{Lavender!20}{Crop}
        & 0.54 & 0.96 & 0.56 & 0.38 & 0.79 & 0.22 & 0.33
        & 20.63 & 5.17 & 0.24 & 0.12 \\

    \multirow{-3}{*}{DiT-I} & \colorbox{Cerulean!20}{CutMix}
        & 0.58 & 0.95 & 0.67 & 0.43 & 0.80 & 0.30 & 0.35
        & 20.81 & 5.34 & 0.25 & 0.26 \\

    \greyrule

    {} & \textcolor{red}{\faTimes}
        & 0.55 & 0.97 & 0.60 & 0.42 & 0.74 & 0.26 & 0.35
        & 20.03 & 5.17 & 0.24 & 0.22 \\

    {} & \colorbox{Lavender!20}{Crop}
        & 0.54 & 0.96 & 0.61 & 0.40 & 0.71 & 0.23 & 0.33
        & 20.29 & 4.72 & 0.23 & -0.1 \\

    \multirow{-3}{*}{TextRIN-I} & \colorbox{Cerulean!20}{CutMix}
        & 0.57 & 0.94 & 0.68 & 0.40 & 0.70 & 0.35 & 0.36
        & 20.03 & 5.16 & 0.24 & 0.30 \\

    \greyrule

    {} & \textcolor{red}{\faTimes}
        & 0.51 & 0.93 & 0.56 & 0.38 & 0.71 & 0.25 & 0.24
        & 19.74 & 4.78 & 0.20 & -0.12 \\

    {} & \colorbox{Lavender!20}{Crop}
        & 0.59 & 0.93 & 0.66 & 0.40 & 0.85 & 0.28 & 0.39
        & 19.91 & 5.15 & 0.22 & 0.07 \\

    \multirow{-3}{*}{SiT-I} & \colorbox{Cerulean!20}{CutMix}
        & 0.61 & 0.96 & 0.69 & 0.47 & 0.80 & 0.34 & 0.40
        & 20.06 & 5.33 & 0.23 & 0.22 \\

    \greyrule

    {} & \textcolor{red}{\faTimes}
        & 0.54 & 0.94 & 0.62 & 0.34 & 0.77 & 0.24 & 0.29
        & 19.79 & 5.06 & 0.22 & -0.01 \\

    {} & \colorbox{Lavender!20}{Crop}
        & 0.51 & 0.95 & 0.53 & 0.33 & 0.74 & 0.20 & 0.26
        & 19.66 & 4.84 & 0.21 & -0.09 \\

    \multirow{-3}{*}{JiT-I} & \colorbox{Cerulean!20}{CutMix}
        & 0.50 & 0.97 & 0.57 & 0.29 & 0.75 & 0.22 & 0.27
        & 19.67 & 4.90 & 0.21 & -0.07 \\

    \bottomrule
 \end{tabular}}
\caption{\textbf{GenEval and Aesthetic metrics} for TA and TA{+}IA models trained only on ImageNet. All models are trained with long captions. A Prompt Extender was used before generating images. Models are evaluated at $256^2$ resolution. Scores are the best obtained within a budget of 400k training steps.}
\label{tab:geneval_plus_aesthetic}
\end{table*}

\paragraph{Results.}
In Figure~\ref{fig:overfitting}, we plot the evolution of GenEval scores over training steps for the TextRIN-I, SiT-I, and JiT-I architectures. Training with TA alone leads to early overfitting for the two latent-space architectures, TextRIN-I and SiT-I. However, adding image augmentation (TA+IA) delays or entirely prevents this overfitting, allowing these latent models to sustain higher GenEval scores for longer. In contrast, image augmentation brings no such benefit to JiT-I. We attribute this difference to the VAE used in latent models: its encoder acts as a perceptual filter that absorbs the sharp, locally unnatural boundary artifacts introduced by CutMix. Consequently, the latent training signal remains close to the natural-image manifold while still benefiting from the added diversity. Because JiT-I models raw pixels directly without this filter, CutMix compromises the visual quality and content the model learns rather than regularizing it.

Table~\ref{tab:geneval_plus_aesthetic} confirms that image augmentations (CutMix, Crop) improve both GenEval and aesthetic metrics for latent-space models, but degrade the pixel-space JiT-I. For DiT-I and TextRIN-I, CutMix boosts overall GenEval by +3 and +2 points, alongside consistent aesthetic gains (e.g., ImageReward +0.08). Sub-task improvements are also notable, such as +9 in Position for TextRIN-I and +2 in Color Attribution for DiT-I. SiT-I sees the most dramatic gains: overall GenEval jumps by +10 points (to 0.61)—driven by Color Attribution (+16) and Two Objects (+13)—while ImageReward surges from -0.12 to 0.22. In contrast, JiT-I metrics decline slightly with either augmentation across both benchmarks (GenEval: CutMix -4, Crop -3)

\begin{figure*}[t]
    \centering
    \renewcommand{\arraystretch}{1.2} 
    \setlength{\tabcolsep}{1.5pt} 
    \begin{tabular}{cccccc}
        \textbf{TA} & \textbf{TA+IA} & \textbf{TA} & \textbf{TA+IA} & \textbf{TA} & \textbf{TA+IA} \\ 

        \includegraphics[width=0.15\textwidth]{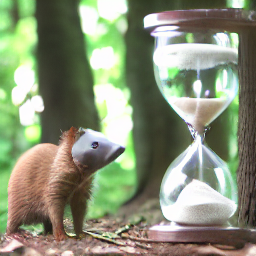} &
        \includegraphics[width=0.15\textwidth]{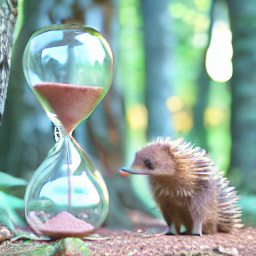} &
        \includegraphics[width=0.15\textwidth]{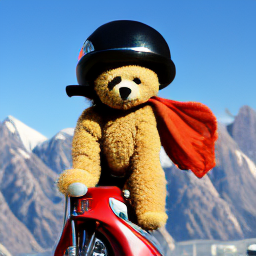} &
        \includegraphics[width=0.15\textwidth]{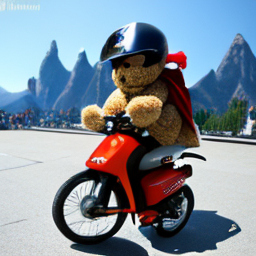} &
        \includegraphics[width=0.15\textwidth]{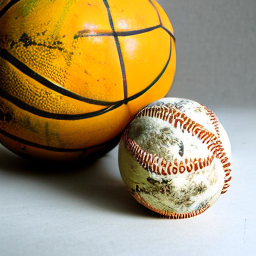} & 
        \includegraphics[width=0.15\textwidth]{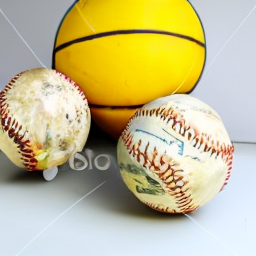} \\
        \multicolumn{2}{c}{\small \texttt{A hedgehog and an hourglass}} &
        \multicolumn{2}{c}{\small \texttt{A teddy bear driving a motorbike}} & \multicolumn{2}{c}{\small \makecell[c]{\texttt{Two beat-up baseballs} \\ \texttt{and a yellow basketball}}} \\[8pt]

        \includegraphics[width=0.15\textwidth]{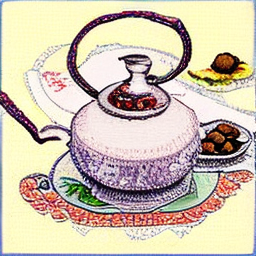} &
        \includegraphics[width=0.15\textwidth]{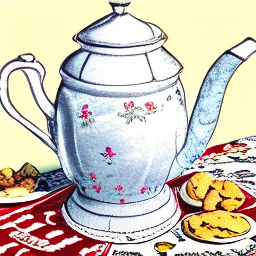} &
        \includegraphics[width=0.15\textwidth]{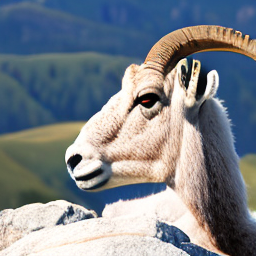} &
        \includegraphics[width=0.15\textwidth]{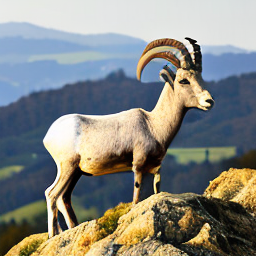} &
        \includegraphics[width=0.15\textwidth]{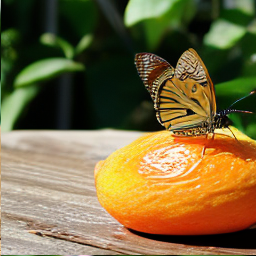} &
        \includegraphics[width=0.15\textwidth]{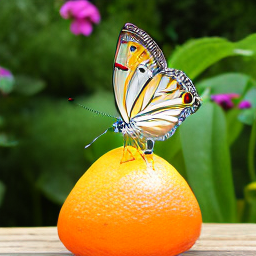} \\
        \multicolumn{2}{c}{\small \texttt{A teapot and cookies on a table}} &
        \multicolumn{2}{c}{\small \texttt{A goat on a mountain top}} & \multicolumn{2}{c}{\small \makecell[c]{\texttt{Multicolored butterfly} \\ \texttt{atop a vibrant tangerine}}} \\[8pt]

    \end{tabular}
    \caption{\textbf{Qualitative comparison: Text-Augmentation (TA, first, third columns) vs Text+Image Augmentation (TA+IA second, last columns)) for four prompts (left and right blocks per row).} Image augmentation improves text comprehension, compositionality  and overall image quality. 
    }
    \label{fig:qualitative_comparison}
\end{figure*}

These improvements in metrics are achieved while maintaining or improving image quality, as measured by FID scores and Prec., Rec., Den., and Cov. in Table \ref{tab:fid_TA_TApIA} (Appendix~\ref{app:quanti}). The qualitative examples in Figure \ref{fig:qualitative_comparison} further demonstrate the enhanced compositional capabilities, with models trained using augmentation techniques producing more accurate representations of complex prompts. This improvement is particularly evident in the teddy bear scene: while the TA model generates a teddy bear awkwardly positioned on a motor bike, the TA {+} IA model creates a more natural composition with the teddy bear appropriately driving the motor bike. Similarly, the ``goat on top of a mountain'' shows refined details and aesthetically pleasing composition with TA {+} IA, whereas the TA model struggles with the scene's layout. The benefits of training with Image Augmentation are also clear in the butterfly and baseball examples, where color attribution is significantly improved. More qualitative examples are given in Figure~\ref{fig:qualitative_comparison_full}.

\subsection{Conditioning with MIRO}
\label{subsec:miro}

MIRO~\citep{dufour2025miro} conditions text-to-image pretraining on multiple reward signals, letting the model learn user preferences directly rather than relying on post-hoc reward fine-tuning. We evaluate MIRO in our ImageNet-only setting on SiT-I. Table~\ref{tab:miro_comparison} compares GenEval and aesthetic metrics without and with MIRO. MIRO conditioning improves both GenEval and aesthetic metrics for SiT-I, with GenEval overall rising by +5 points and ImageReward improving from $-$0.12 to $0.29$. However, the improvement brought by MIRO in our ImageNet-only setting is smaller than that reported in the original paper on its CC12M-LAION-$6+$ dataset, and remains weaker than what CutMix alone achieves in Table~\ref{tab:geneval_plus_aesthetic}. Figure~\ref{fig:miro_progression_single} shows visually the speed-up and the improvement brought by the MIRO conditioning.

\begin{table*}[]
\centering
\resizebox{\linewidth}{!}{
\begin{tabular}{l c @{\;}!{\color{gray!60}\vrule} c c c c c c c @{\;}!{\color{gray!60}\vrule} c c c c}
    \toprule

    \multirow{2}{*}{\textbf{Model}} &
    \multirow{2}{*}{\textbf{MIRO}} &
    \multicolumn{7}{c!{\color{gray!60}\vrule}}{\textbf{GenEval}} &
    \multicolumn{4}{c}{\textbf{Aesthetic metrics}} \\

    \cmidrule(lr){3-9} \cmidrule(lr){10-13}

    &
    & \textbf{Overall$\uparrow$} & \textbf{One obj.$\uparrow$} & \textbf{Two obj.$\uparrow$} & \textbf{Count.$\uparrow$} & \textbf{Col.$\uparrow$} & \textbf{Pos.$\uparrow$} & \textbf{Col. attr.$\uparrow$}
    & \textbf{PickScore$\uparrow$} & \textbf{Aes.Score$\uparrow$} & \textbf{HPSv2.1$\uparrow$} & \textbf{ImageReward$\uparrow$} \\

    \midrule

    {} & \textcolor{red}{\faTimes}
        & 0.51 & 0.93 & 0.56 & 0.38 & 0.71 & 0.25 & 0.24
        & 19.74 & 4.78 & 0.20 & -0.12 \\

    \multirow{-2}{*}{SiT-I L} & \faCheck
        & 0.56 & 0.95 & 0.63 & 0.46 & 0.71 & 0.36 & 0.25
        & 20.10 & 5.10 & 0.23 & 0.29 \\

    \bottomrule
 \end{tabular}}
\caption{\textbf{SiT-I with MIRO}~\citep{dufour2025miro}. GenEval and Aesthetic metrics for SiT-I L without and with MIRO conditioning (with TA but without IA).}
\label{tab:miro_comparison}
\end{table*}

\input{images/miro_imagenet_visual_progression}

\subsection{Scaling to Bigger Models}
\label{subsec:scaling_bigger_models}

We investigate whether performance keeps increasing with larger model sizes. We take the best-performing configuration (SiT-I with CutMix) and scale it up from L to XL and H. Table~\ref{tab:scaling_bigger_models} reports GenEval and aesthetic metrics across scales. Compositionality does not improve with model size: the GenEval overall score stays flat or even slightly decreases when moving from L to XL/H. We hypothesize that most of the compositional capacity attainable in our ImageNet-only setting is already extracted by the L model, leaving little room for larger models to improve further on GenEval. Aesthetic metrics, in contrast, improve slightly with scale: Aesthetics Score and ImageReward both increase from L to XL/H.

\begin{table*}[]
\centering
\resizebox{\linewidth}{!}{
\begin{tabular}{l c @{\;}!{\color{gray!60}\vrule} c c c c c c c @{\;}!{\color{gray!60}\vrule} c c c c}
    \toprule

    \multirow{2}{*}{\textbf{Model}} &
    \multirow{2}{*}{\textbf{Size}} &
    \multicolumn{7}{c!{\color{gray!60}\vrule}}{\textbf{GenEval}} &
    \multicolumn{4}{c}{\textbf{Aesthetic metrics}} \\

    \cmidrule(lr){3-9} \cmidrule(lr){10-13}

    &
    & \textbf{Overall$\uparrow$} & \textbf{One obj.$\uparrow$} & \textbf{Two obj.$\uparrow$} & \textbf{Count.$\uparrow$} & \textbf{Col.$\uparrow$} & \textbf{Pos.$\uparrow$} & \textbf{Col. attr.$\uparrow$}
    & \textbf{PickScore$\uparrow$} & \textbf{Aes.Score$\uparrow$} & \textbf{HPSv2.1$\uparrow$} & \textbf{ImageReward$\uparrow$} \\

    \midrule

    {} & L
        & 0.61 & 0.96 & 0.69 & 0.47 & 0.80 & 0.34 & 0.40
        & 20.06 & 5.33 & 0.23 & 0.22 \\

    {} & XL
        & 0.58 & 0.95 & 0.69 & 0.46 & 0.79 & 0.27 & 0.34
        & 20.13 & 5.33 & 0.23 & 0.23 \\

    \multirow{-3}{*}{\makecell{SiT-I\\\colorbox{Cerulean!20}{CutMix}}} & H
        & 0.61 & 0.97 & 0.70 & 0.44 & 0.81 & 0.30 & 0.41
        & 20.14 & 5.37 & 0.23 & 0.30 \\

    \bottomrule
 \end{tabular}}
\caption{\textbf{Scaling to bigger models.} GenEval and Aesthetic metrics for SiT-I L, XL and H, without MIRO.}
\label{tab:scaling_bigger_models}
\end{table*}

\subsection{Scaling to Higher Resolution}
\label{subsec:higher_resolution}

All experiments above use $256^2$ resolution. We now explore whether higher-resolution generation is feasible under the same data constraints, and investigate two approaches to scale to $512^2$: (i) fine-tuning an existing $256^2$ checkpoint, and (ii) training a model from scratch directly at $512^2$.

\vspace{2mm} \noindent \textbf{Fine-tuning from a $256^2$ checkpoint.} Starting from a DiT-I checkpoint at $256^2$ trained with CutMix, we further train it at $512^2$ for 50k steps. This fine-tuning requires no changes except to adjust the image tokenization. Table~\ref{tab:geneval_aesthetic_resolution} reports GenEval and aesthetic metrics across resolutions. Higher-resolution fine-tuning improves the GenEval score from \textbf{0.58} at $256^2$ to \textbf{0.61} at $512^2$. Images at $512^2$ are shown in Appendix~\ref{app:quali_512}.

\vspace{2mm} \noindent \textbf{Training from scratch at $512^2$.} We train from scratch a SiT-I model at $512 \times 512$ resolution. Similarly, increasing the resolution improves both GenEval (from $0.61$ to $0.63$ overall score) and the Aesthetic metrics. 

\begin{table*}[ht]
    \centering
    \resizebox{\linewidth}{!}{
    \begin{tabular}{l c !{\color{gray!60}\vrule} c @{\;}!{\color{gray!60}\vrule} c c c c c c @{\;}!{\color{gray!60}\vrule} c c c c}
        \toprule
        \multirow{2}{*}{\textbf{Model}} & \multirow{2}{*}{\textbf{Res.}} & \multicolumn{7}{c!{\color{gray!60}\vrule}}{\textbf{GenEval}} & \multicolumn{4}{c}{\textbf{Aesthetic metrics}} \\
        \cmidrule(lr){3-9} \cmidrule(lr){10-13}
        & & \textbf{Overall$\uparrow$} & \textbf{One obj.$\uparrow$} & \textbf{Two obj.$\uparrow$} & \textbf{Count.$\uparrow$} & \textbf{Col.$\uparrow$} & \textbf{Pos.$\uparrow$} &
        \textbf{Col. attr.$\uparrow$}
        & \textbf{PickScore$\uparrow$} & \textbf{Aes.\ Score$\uparrow$} & \textbf{HPSv2.1$\uparrow$} & \textbf{ImageReward$\uparrow$} \\
        \midrule
        \multicolumn{13}{l}{\textit{Finetuning}} \\
        {} & $256^2$ & 0.58 & 0.95 & 0.67 & 0.43 & 0.80 & 0.30 & 0.35
        & 20.81 & 5.34 & 0.25 & 0.26 \\
        \multirow{-2}{*}{DiT-I \colorbox{Cerulean!20}{CutMix}} & $512^2$ & 0.61 & 0.98  & 0.73 & 0.43 & 0.76 & 0.34 & 0.40
        & 21.05 & 5.60 & 0.25 & 0.40 \\
        \midrule
        \multicolumn{13}{l}{\textit{From scratch}} \\
        {} & $256^2$ & 0.61 & 0.96 & 0.69 & 0.47 & 0.80 & 0.34 & 0.40
        & 20.06 & 5.33 & 0.23 & 0.22 \\
        \multirow{-2}{*}{SiT-I \colorbox{Cerulean!20}{CutMix}} & $512^2$ & 0.63 & 0.97 & 0.76 & 0.46 & 0.81 & 0.32 & 0.45
        & 20.46 & 5.66 & 0.24 & 0.37 \\
        \bottomrule
    \end{tabular}
    }
    \caption{\textbf{GenEval and Aesthetic metrics} across resolutions for DiT-I  and SiT-I.}
    \label{tab:geneval_aesthetic_resolution}
\end{table*}

\subsection{Putting Everything Together}
\label{subsec:putting_together}

We combine the best ingredients identified above into a single model: SiT-H trained with CutMix and MIRO conditioning. Table~\ref{tab:putting_together} reports GenEval and aesthetic metrics for this model at $256^2$ and $512^2$ resolution. We achieve the best results with this model: \textbf{0.67} GenEval overall score and \textbf{0.27} HPSv2 score.

\begin{table*}[]
\centering
\resizebox{\linewidth}{!}{
\begin{tabular}{l c @{\;}!{\color{gray!60}\vrule} c c c c c c c @{\;}!{\color{gray!60}\vrule} c c c c}
    \toprule

    \multirow{2}{*}{\textbf{Model}} &
    \multirow{2}{*}{\textbf{Res.}} &
    \multicolumn{7}{c!{\color{gray!60}\vrule}}{\textbf{GenEval}} &
    \multicolumn{4}{c}{\textbf{Aesthetic metrics}} \\

    \cmidrule(lr){3-9} \cmidrule(lr){10-13}

    &
    & \textbf{Overall$\uparrow$} & \textbf{One obj.$\uparrow$} & \textbf{Two obj.$\uparrow$} & \textbf{Count.$\uparrow$} & \textbf{Col.$\uparrow$} & \textbf{Pos.$\uparrow$} & \textbf{Col. attr.$\uparrow$}
    & \textbf{PickScore$\uparrow$} & \textbf{Aes.Score$\uparrow$} & \textbf{HPSv2.1$\uparrow$} & \textbf{ImageReward$\uparrow$} \\

    \midrule

    {} & $256^2$
        & 0.65 & 0.98 & 0.73 & 0.50 & 0.81 & 0.38 & 0.49
        & 20.59 & 5.55 & 0.27 & 0.67 \\

    \multirow{-2}{*}{SiT-H} & $512^2$
        & 0.67 & 0.98 & 0.82 & 0.47 & 0.85 & 0.36 & 0.52
        & 20.84 & 5.77 & 0.28 & 0.73 \\

    \bottomrule
 \end{tabular}}
\caption{\textbf{Putting everything together.} GenEval and Aesthetic metrics for SiT-H trained on ImageNet only with CutMix and MIRO conditioning at $256^2$ and $512^2$ resolution.}
\label{tab:putting_together}
\end{table*}


\section{State-of-the-art comparison}
\label{sec:comp_sota}

We investigate how good a model trained solely on ImageNet fares against the state of the art. We compare the SiT-I H trained at $512^2$ resolution with CutMix and MIRO conditioning from Section~\ref{subsec:putting_together}.






\vspace{2mm} \noindent \textbf{GenEval.}
Table \ref{tab:sota_geneval} reports the results on GenEval benchmark. Our best model reaches an overall score of \textbf{0.67}, matching FLUX-dev (\textbf{0.67}) and surpassing SANA-0.6 (\textbf{0.64}) and Stable Diffusion 3 (SD3 M - \textbf{0.62}), both trained on orders of magnitude more data. It reaches the best score on Two Objects (\textbf{0.82}) and Color Attribution (\textbf{0.52}), but falls behind on Counting (\textbf{0.47} compated to \textbf{0.63} for SANA-0.6 and SD3). 



\begingroup

\begin{table*}[t]
    \centering
    \resizebox{\linewidth}{!}{
        \begin{tabular}{l@{\;}c@{\;}c@{\;}!{\color{gray!60}\vrule}c@{\;\;}!{\color{gray!60}\vrule}c@{\;\;}c@{\;\;}c@{\;\;}c@{\;\;}c@{\;\;}c@{\;\;}c@{\;}}
            \toprule
             \textbf{Model} & \textbf{Size} & \makecell[c]{\textbf{\#data}} & \textbf{Overall$\uparrow$} & \makecell[c]{\textbf{One obj.}}$\uparrow$ & \makecell[c]{\textbf{Two obj.}}$\uparrow$ & \textbf{Coun. $\uparrow$} & \textbf{Col. $\uparrow$} & \textbf{Pos. $\uparrow$} & \makecell[c]{\textbf{Col. Attr.}}$\uparrow$ \\
            \midrule
    
            SD v1.5~\cite{rombach2022high} & 0.9B & 5B+ & 0.43 & 0.97 & 0.38 & 0.35 & 0.76 & 0.04 & 0.06 \\
            PixArt-$\alpha$~\cite{chen2023pixart}  & 0.6B & 25M & 0.48 & \underline{0.98} & 0.50 & 0.44 & 0.80 & 0.08 & 0.07 \\
            PixArt-$\Sigma$ (512)~\cite{chen2024pixart} & 0.6B & 35M+ & 0.52 & \underline{0.98} & 0.59 & 0.50 & 0.80 & 0.10 & 0.15 \\
            SD v2.1~\cite{rombach2022high} & 0.9B & 5B+ & 0.50 & \underline{0.98} & 0.51 & 0.44 & \underline{0.85} & 0.07 & 0.17 \\
            SDXL~\cite{podell2023sdxl} & 3.5B & 5B+  & 0.55 & \underline{0.98} & 0.74 & 0.39 & \underline{0.85} & 0.15 & 0.23 \\
             SD3 M (512)~\cite{esser2024scaling} & 2B & 1B+ & 0.62 & \underline{0.98} & 0.74 & 0.63 & 0.67 & 0.34 & 0.36 \\
             SANA-0.6~\cite{xie2024sana} & 0.6B & \textcolor{orange}{\faBan} & 0.64 & \textbf{0.99} & 0.71 & 0.63 & \textbf{0.91} & 0.16 & 0.42 \\
             FLUX-dev~\cite{labs2025flux1kontextflowmatching} & 12B &\textcolor{orange}{\faBan} & \underline{0.67} & \textbf{0.99} & \underline{0.81} & \textbf{0.79} & 0.74 & 0.20 & 0.47 \\
            MIRO~\cite{dufour2025miro} & 0.3B & 18M & \textbf{0.75} & \underline{0.98} & 0.79 & \underline{0.71} & \underline{0.85} & \textbf{0.58} & \textbf{0.58} \\ 
            \greyrule
             \rowcolor{Periwinkle!10} Ours (SiT-I H CutMix MIRO) & 1B & 1.2M & \underline{0.67} & \underline{0.98} & \textbf{0.82} & 0.47 & \underline{0.85} & \underline{0.36} & \underline{0.52} \\  
            \bottomrule
        \end{tabular}}
    \caption{\textbf{Results on GenEval}. \textbf{Bold} indicates best, \underline{underline} second best.}  
    \label{tab:sota_geneval}
    
\end{table*}

\endgroup

\begin{table}[]
    \centering
    \small
    \begin{tabular}{l@{\;}c@{\;}!{\color{gray!60}\vrule}c@{\;}c@{\;}c@{\;}c@{\;}c@{\;}}
        \toprule
        \textbf{Model}   & \textbf{Overall$\uparrow$} & \textbf{Entity$\uparrow$} & \textbf{Attribute$\uparrow$} & \textbf{Relation$\uparrow$} & \textbf{Other$\uparrow$} & \textbf{Global$\uparrow$} \\
        \midrule
        SDv1.5~\cite{rombach2022high}  & 63.2 & 74.2 & 75.4 & 73.5 & 67.8 & 74.6 \\
        Pixart-$\alpha$~\cite{chen2023pixart} & 71.1 & 79.3 & 78.6 & 82.6 & 77.0 &  75.0\\
        CAD~\cite{dufour2024don} & 77.6 & 85.3 & 84.7 & \underline{91.5} & 74.8 & 84.5 \\
        Pixart-$\Sigma$ (512)~\cite{chen2024pixart} & 79.5 & 87.1 & 86.5 & 84.0 & 86.1 & \underline{87.5} \\
        Pixart-$\Sigma$ (1024)~\cite{chen2024pixart}  & 80.5 & 82.9 & \textbf{88.9} & 86.6 & 87.7 & 86.9 \\
        Sana-0.6~\cite{xie2024sana} & \textbf{84.3} & \underline{90.0} & 88.6 & 90.1 & \textbf{91.9} & 82.6 \\ 
        SDXL~\cite{podell2023sdxl}  & 74.7 & 82.4 & 80.9 & 86.8 & 80.4 & 83.3 \\
        SD3-Medium~\cite{esser2024scaling}  & \underline{84.1}  & \textbf{91.0} & \underline{88.8} & 80.7 & 88.7 &  \textbf{87.9} \\
        Janus~\cite{wu2024janus}  & 79.7 & 87.4 & 87.7 & 85.5 & 86.4 &  82.3 \\
        FLUX-dev~\cite{labs2025flux1kontextflowmatching} & 84.0 & 89.5 & 88.7 & {91.1} & \underline{89.4} &  82.1 \\
        \greyrule
        \rowcolor{Periwinkle!10} Ours (SiT-I H CutMix MIRO) & 81.2 & 88.4 & 86.4 & \textbf{93.0} & 76.0 & 82.1 \\ 
        \bottomrule \\
    \end{tabular}
    \caption{\textbf{Results on DPG-Bench}. We compare our models to the results reported in~\cite{wu2024janus}. \textbf{Bold} indicates best, \underline{underline} second best. }
    \label{tab:dpgbench}
\end{table}


\vspace{2mm} \noindent \textbf{DPGBench.}
Table \ref{tab:dpgbench} presents results on DPGBench, a more complex benchmark than GenEval, showing consistent trends. Our model attains \textbf{81.2\%} overall accuracy, surpassing SDXL by 6 pp and PixArt-$\Sigma$ by 2 pp, and matching the performance of Janus. It trails Flux-dev and SANA-0.6 by 3 pp but achieves state of the art on \textit{Relation} with \textbf{93\%}.




\section{Generalisation Analysis}
\label{sec:generalisation}


ImageNet comes with an object centric bias. One could believe that models trained on ImageNet would exhibit similar characteristics, which goes against the desired expectation of generic T2I models, i.e. generating complex scenes or interactions. Quantitatively, Tables \ref{tab:sota_geneval} and \ref{tab:dpgbench} indicate that our models do hold strong for complex scenes by having competitive scores in many subcategories of GenEval and DPG like Relation, Position, Two Object, and Color Attribution. 

\begin{figure*}
    \centering
    \renewcommand{\arraystretch}{1.2} 
    \setlength{\tabcolsep}{3pt} 
    \begin{tabular}{cccc}


         \includegraphics[width=0.23\textwidth]{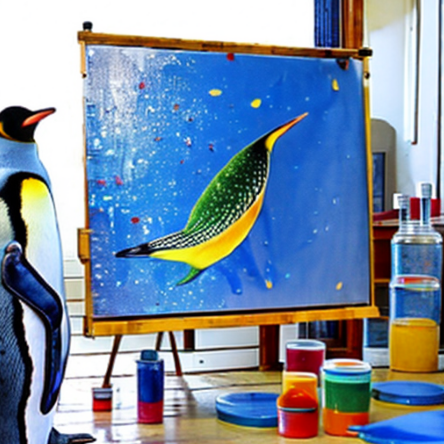} &

         \includegraphics[width=0.23\textwidth]{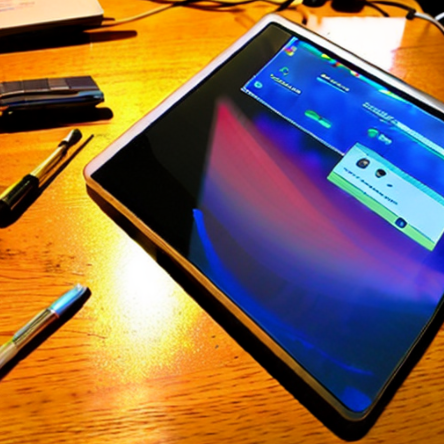} &

         \includegraphics[width=0.23\textwidth]{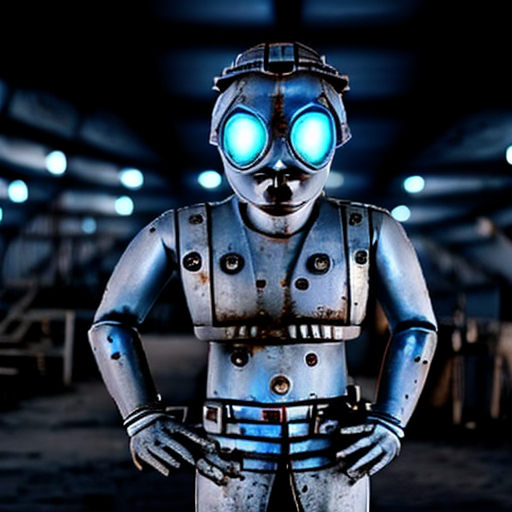} &

         \includegraphics[width=0.23\textwidth]{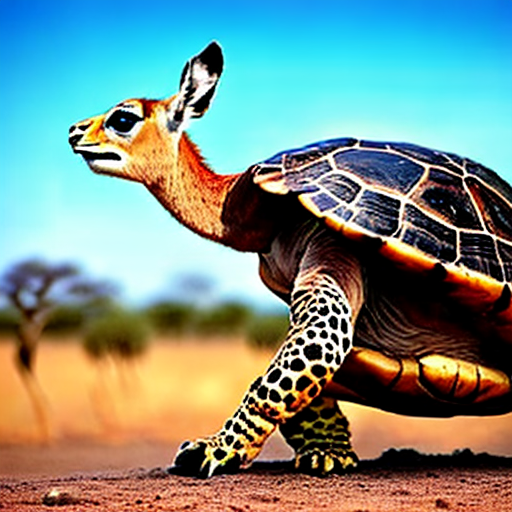} \\

         {\parbox{0.22\textwidth}{\tiny\texttt{A penguin stands in a small art studio holding a paintbrush dipped in blue paint, working on a canvas mounted.}}} & 
         
         {\parbox{0.22\textwidth}{\tiny\texttt{An iPad Pro rests on a rustic wooden table. The tablet is positioned at a slight angle.}}} & 

         {\parbox{0.22\textwidth}{\tiny\texttt{A metallic robot standing with faint blue light eyes}}} & 
         
         {\parbox{0.22\textwidth}{\tiny\texttt{A hybrid creature with the body and shell of a tortoise and the head and neck of a giraffe.}}} \\[8pt]

    \end{tabular}
    \caption{\textbf{Generalization examples}. We show that models trained only on ImageNet are capable of producing compositional images with complex scenes and novel concepts. Left images have been generated without MIRO conditioning, right images with.} 
    \label{fig:vis_gen}
\end{figure*}

Qualitatively, when prompted with highly out-of-distribution prompts, our models generate visually pleasing images with a notable T2I alignment, as shown in Figure~\ref{fig:vis_gen}. The model is able to approximate the appearance of concepts that it has not seen during training with patterns related to training concepts. For example, it generates an hybrid creature by combining elements of a giraffe and a turtle, by leveraging the text understanding of the LLM producing the text tokens.


CutMix training also introduces its own failure modes on semantically disjoint prompts, along with a learned mitigation at higher resolution; see Appendix~\ref{subsec:failures} for a full analysis.

\section{Conclusion}
\label{sec:discussion}


We challenged the prevailing wisdom that billion-scale datasets are necessary to unlock text-to-image generation and suggest that this is merely a sufficient condition.
These large-scale datasets are usually either closed-sourced or rapidly decaying, which threatens openness and reproducibility in text-to-image generation research.
Instead, we show that it is possible to train smaller models to high quality using ImageNet only yielding general text-to-image capabilities.

We propose a standardized text-to-image training setup on ImageNet that leads to models capable of generating high quality images while being excellent at prompt following. This is attested by results on common benchmarks matching or outclassing company models widely recognized as good text-to-image generators such as FLUX-dev \cite{labs2025flux1kontextflowmatching}, or the Stable Diffusion family \citep{esser2024scaling, podell2023sdxl, rombach2022high}.

The implications of our work extend beyond just computational efficiency, open science and reproducibility. By showing that smaller datasets achieve state-of-the-art results, we open \textbf{new possibilities for specialized domains}, where large-scale data collection is impractical. Our work also suggests a path toward more controllable and ethical development of text-to-image models, as smaller datasets enable more thorough content verification and bias mitigation.


Looking forward, we believe our results will encourage the community to \textbf{reconsider the ``bigger is better''} paradigm. Future work could explore additional augmentation strategies, investigate the theoretical foundations of data efficiency, and develop even more compact architectures optimized for smaller datasets. Ultimately, we hope this work starts a shift toward more \textbf{sustainable and responsible development }of text-to-image generation models.

More broadly, this work offers a \textbf{complementary perspective} to large-scale T2I models. ImageNet provides a stable, open, and controlled setting that enables reproducible experimentation and clearer analysis of data. We hope this encourages the community to pursue principled, standardized evaluations alongside continued progress in scale.

\newpage

\section*{Acknowledgment}
This work was granted access to the HPC resources of IDRIS under the allocations 2025-AD011015436, 2025-AD011015594 and 2026-A0201017545 made by GENCI. This work was supported by a Hi!Paris grant, ANR/France 2030 program (ANR-23-IACL-0005) and ANR project sharp ANR-23-
PEIA-0008 in the context of the PEPR IA. The authors would like to thank Alexei A. Efros, Thibaut Loiseau, Yannis Siglidis, Yohann Perron, Louis Geist, Robin Courant and Sinisa Stekovic for their insightful comments, suggestions, and discussions.

\bibliography{shortstrings,bibliography}
\bibliographystyle{tmlr}

\newpage
\appendix


\section{Additional Qualitative Results}
\label{app:quali_512}

\subsection{Qualitative Results at 512\textsuperscript{2}}
\label{app:quali_sota_512}

\begin{figure*}[p]
    \centering
    \setlength{\tabcolsep}{2pt} 
    \renewcommand{\arraystretch}{0} 
    \begin{tabular}{cccc}
        \includegraphics[width=0.24\textwidth]{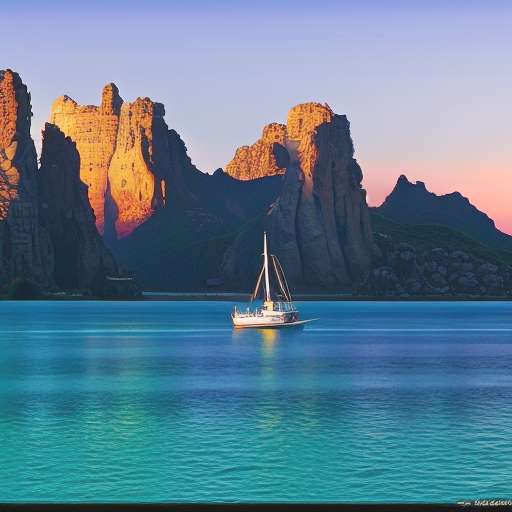} &
        \includegraphics[width=0.24\textwidth]{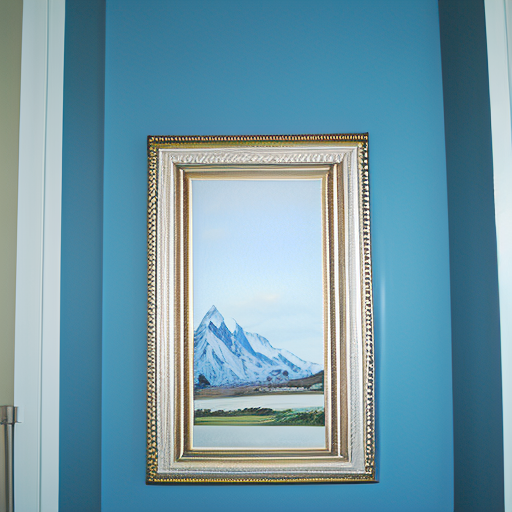} &
        \includegraphics[width=0.24\textwidth]{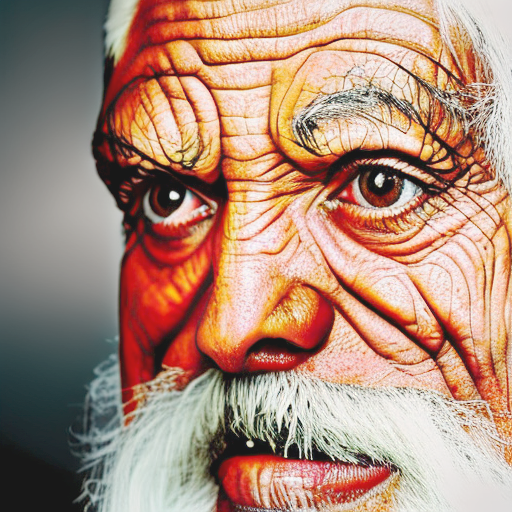} &
        \includegraphics[width=0.24\textwidth]{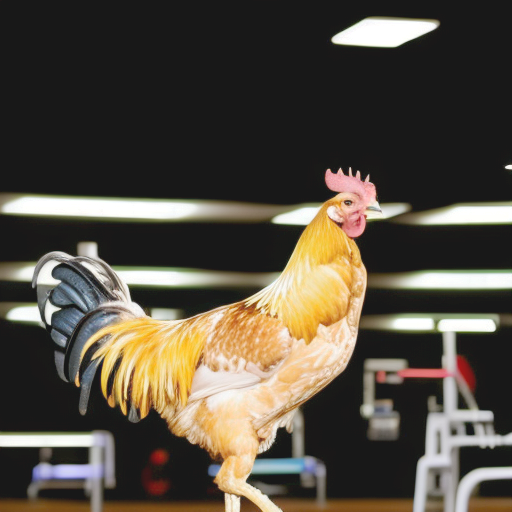} \\ \noalign{\vspace{5pt}}
        \includegraphics[width=0.24\textwidth]{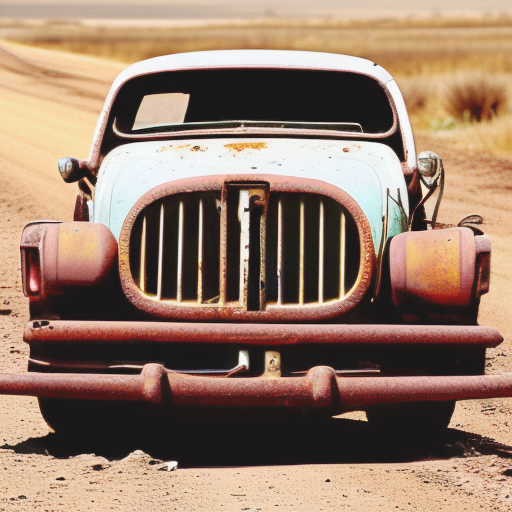} &
        \includegraphics[width=0.24\textwidth]{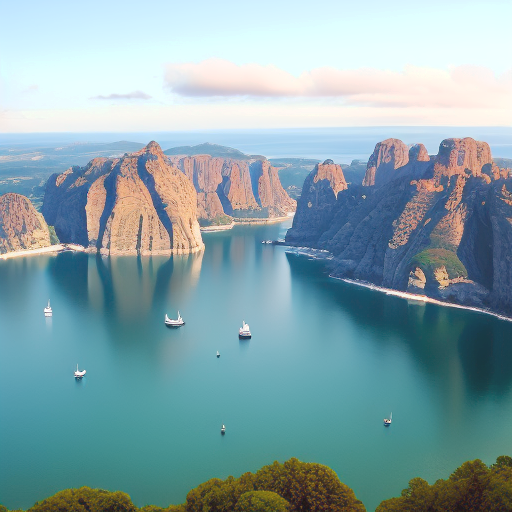} &
        \includegraphics[width=0.24\textwidth]{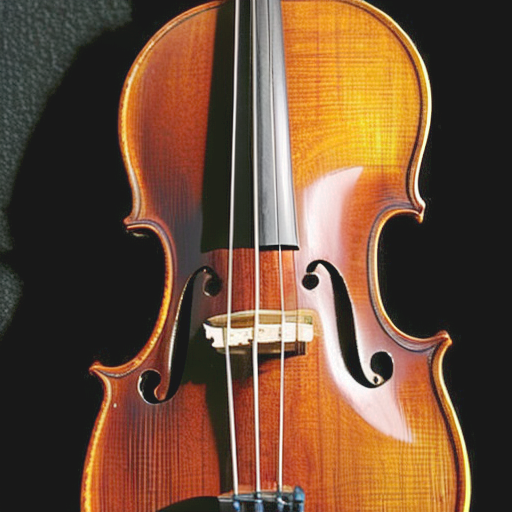} &
        \includegraphics[width=0.24\textwidth]{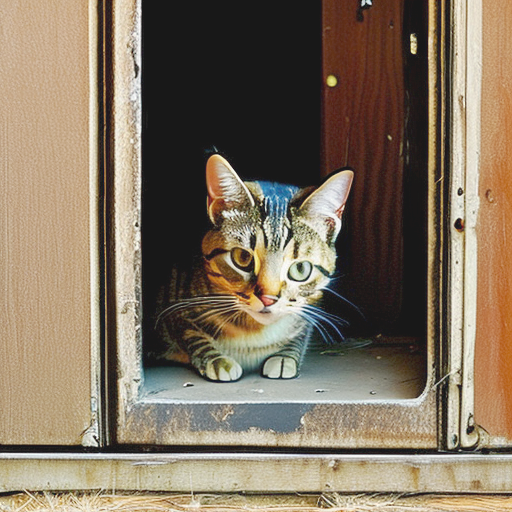} \\ \noalign{\vspace{5pt}}
        \includegraphics[width=0.24\textwidth]{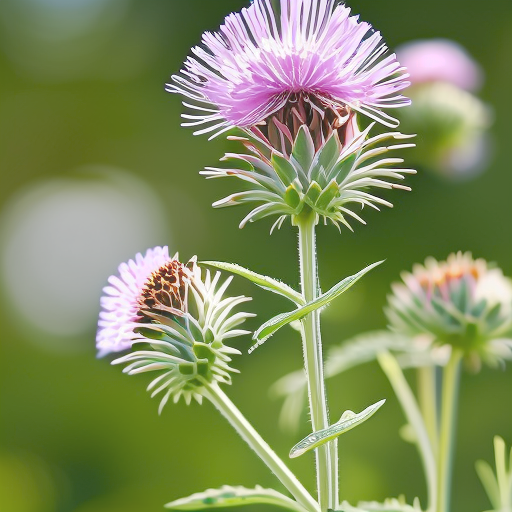} &
        \includegraphics[width=0.24\textwidth]{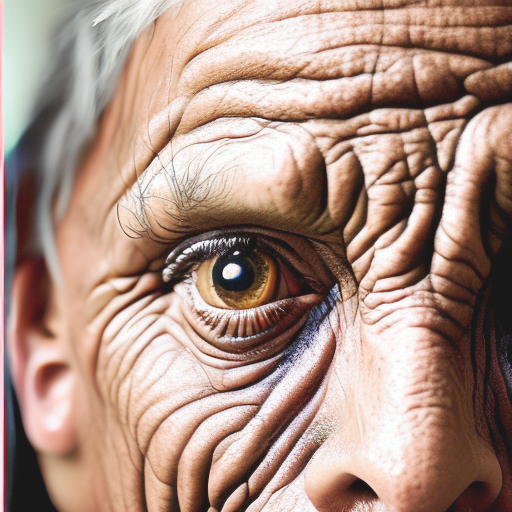} &
        \includegraphics[width=0.24\textwidth]{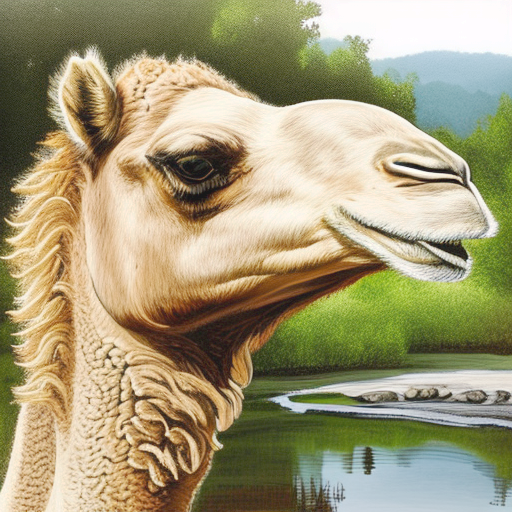} &
        \includegraphics[width=0.24\textwidth]{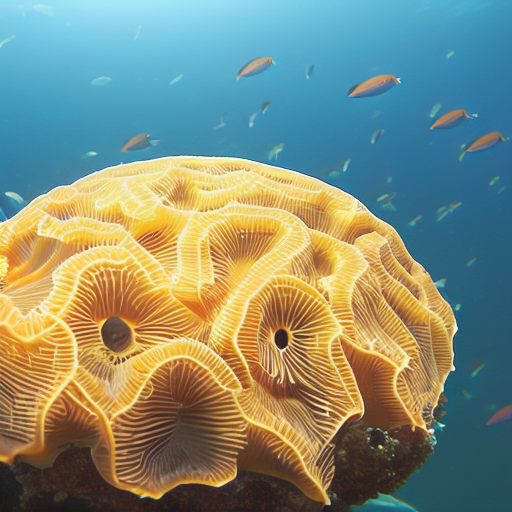} \\ \noalign{\vspace{5pt}}
        \includegraphics[width=0.24\textwidth]{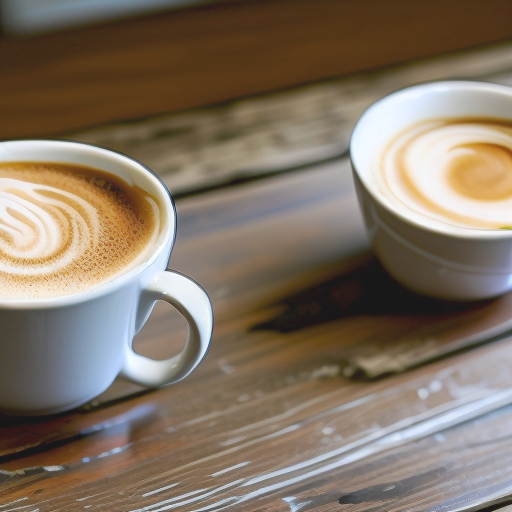} &
        \includegraphics[width=0.24\textwidth]{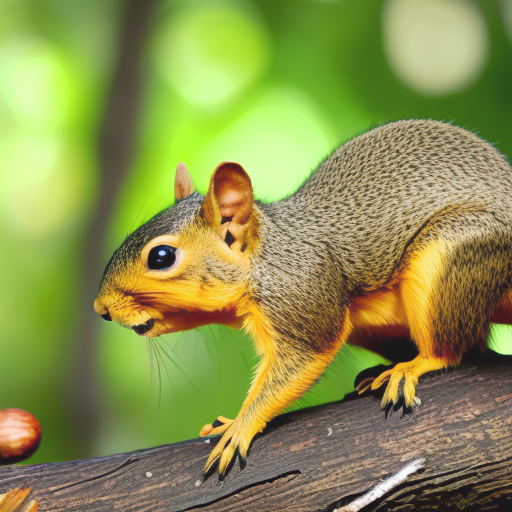} &
        \includegraphics[width=0.24\textwidth]{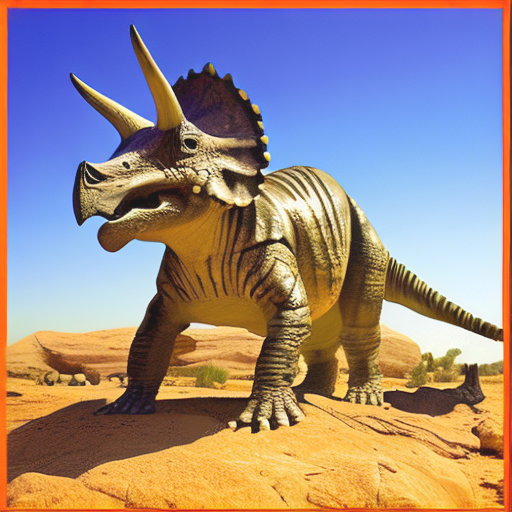} &
        \includegraphics[width=0.24\textwidth]{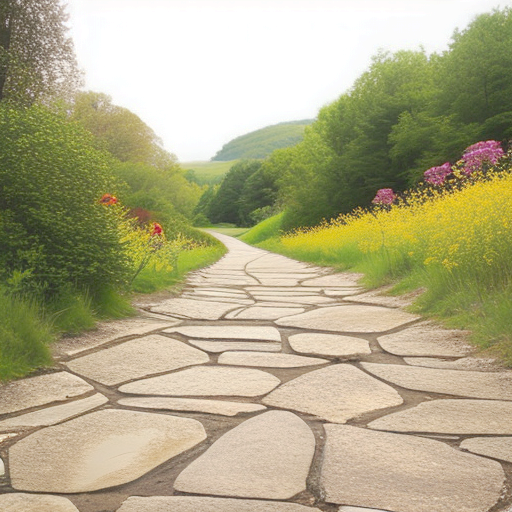} \\ \noalign{\vspace{5pt}}
        \includegraphics[width=0.24\textwidth]{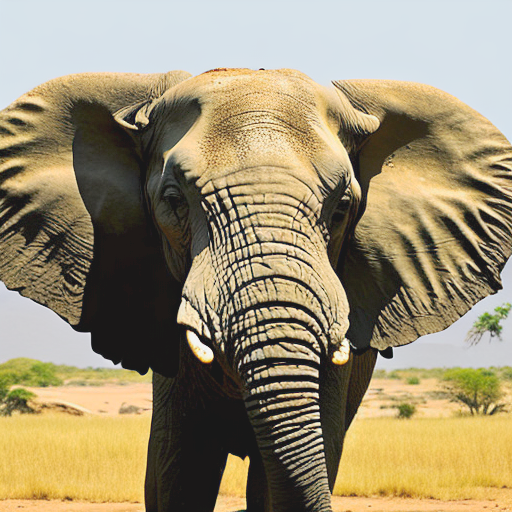} &
        \includegraphics[width=0.24\textwidth]{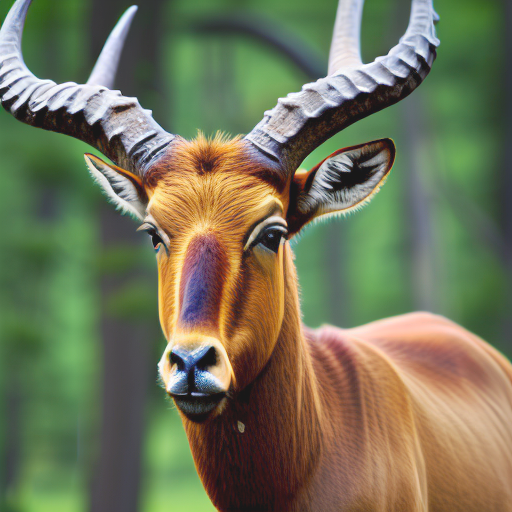} &
        \includegraphics[width=0.24\textwidth]{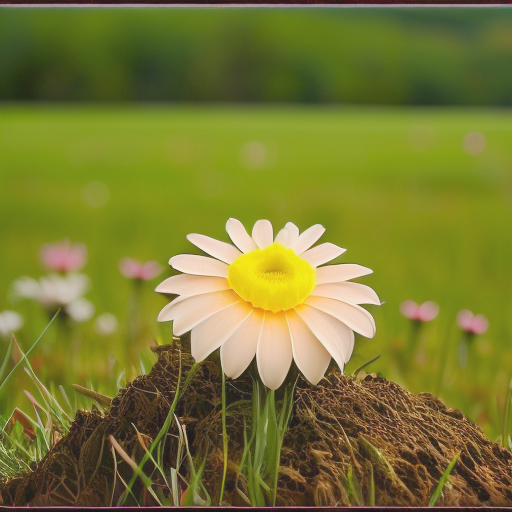} &
        \includegraphics[width=0.24\textwidth]{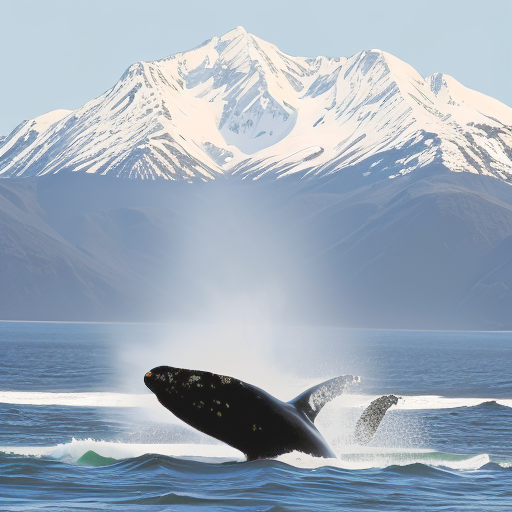} \\ \noalign{\vspace{5pt}}  
    \end{tabular}
    \caption{Additional Qualitative Results at $512^2$ resolution}
    \label{fig:image_grid_512}
\end{figure*}

Figure~\ref{fig:image_grid_512} presents extended qualitative samples generated by our model DiT-I trained with CutMix at $512\times 512$ resolution (see Section \ref{subsec:higher_resolution}).

\subsection{Qualitative Ablation: AIO vs.\ TA vs.\ TA+IA}
\label{app:quali_abla}

\begin{figure*}[p]
    \centering
    \renewcommand{\arraystretch}{0.8} 
    \setlength{\tabcolsep}{3pt} 
    \begin{tabular}{ccc}
        \textbf{AIO} & \textbf{TA} & \textbf{TA+IA} \\ 

        \includegraphics[width=0.21\textwidth]{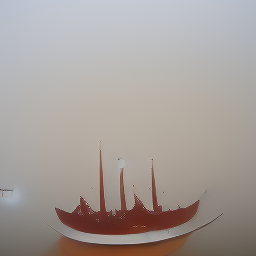} &
        \includegraphics[width=0.21\textwidth]{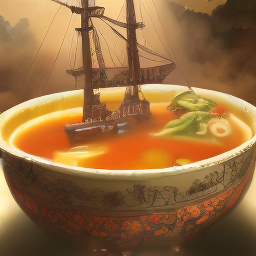} &
        \includegraphics[width=0.21\textwidth]{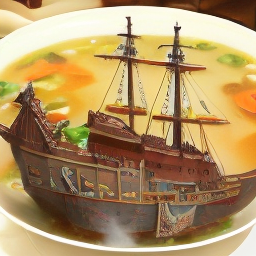} \\
        \multicolumn{3}{c}{\small \texttt{A pirate ship sailing on a steaming soup}} \\[10pt]

        \includegraphics[width=0.21\textwidth]{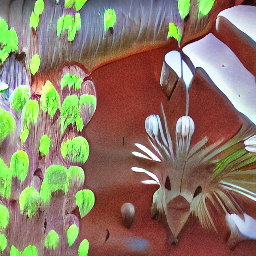} &
        \includegraphics[width=0.21\textwidth]{images/fig_qualitative_comparison/ta_hedeghog.png} &
        \includegraphics[width=0.21\textwidth]{images/fig_qualitative_comparison/ia_hedgedog.png} \\
        \multicolumn{3}{c}{\small \texttt{A hedgehog and an hourglass}} \\[10pt]

        \includegraphics[width=0.21\textwidth]{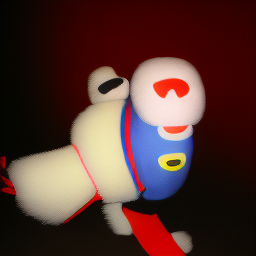} &
        \includegraphics[width=0.21\textwidth]{images/fig_qualitative_comparison/ta_teddybear.png} &
        \includegraphics[width=0.21\textwidth]{images/fig_qualitative_comparison/ia_teddybear.png} \\
        \multicolumn{3}{c}{\small \texttt{A teddy bear driving a motorbike}} \\[10pt]

        \includegraphics[width=0.21\textwidth]{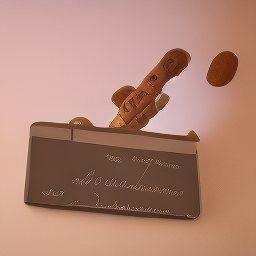} &
        \includegraphics[width=0.21\textwidth]{images/fig_qualitative_comparison/ta_tea.png} &
        \includegraphics[width=0.21\textwidth]{images/fig_qualitative_comparison/ia_tea.png} \\
        \multicolumn{3}{c}{\small \texttt{A teapot and cookies on a table}} \\[10pt]

        \includegraphics[width=0.21\textwidth]{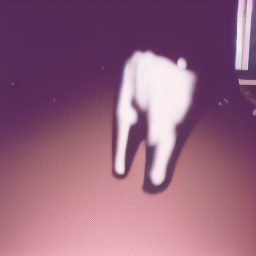} &
        \includegraphics[width=0.21\textwidth]{images/fig_qualitative_comparison/ta_goat.png} &
        \includegraphics[width=0.21\textwidth]{images/fig_qualitative_comparison/ia_goat.png} \\
        \multicolumn{3}{c}{\small \texttt{A goat on a mountain top}} \\[10pt]

    \end{tabular}
    \vspace{-0.3cm}
    \caption{\textbf{Qualitative comparison across models: AIO, TA, and TA+IA}. Image and text augmentations improve text comprehension and overall image quality.}
    \label{fig:qualitative_comparison_full}
\end{figure*}

Figure~\ref{fig:qualitative_comparison_full} shows the influence of the Text (TA) and Image (TA+IA) augmentations against the baseline (AIO). The examples demonstrate how combining image and text augmentations improves semantic alignment, object relations, and overall scene realism. See Section \ref{subsec:images_aug}. 

\section{CutMix Failure Cases and Mitigations}
\label{subsec:failures}

\begin{figure*}[!h]
    \centering
    \renewcommand{\arraystretch}{1.2}
    \setlength{\tabcolsep}{3pt}

    \begin{tabular}{c}
        \makecell[c]{\textbf{Prompt: The image depicts a juxtaposition where the top part shows an indoor} \\ \textbf{setting with a piece of furniture [\ldots]. The bottom part of the image presents an outdoor} \\ \textbf{scene with a focus on a red, rounded object that [\ldots]}} \\[10pt]

        \includegraphics[width=0.6\linewidth]{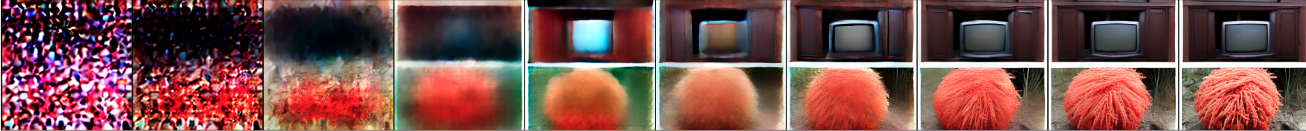} \\[5pt]
        \small \texttt{Model trained without CutMix} \\[10pt]

        \includegraphics[width=0.6\linewidth]{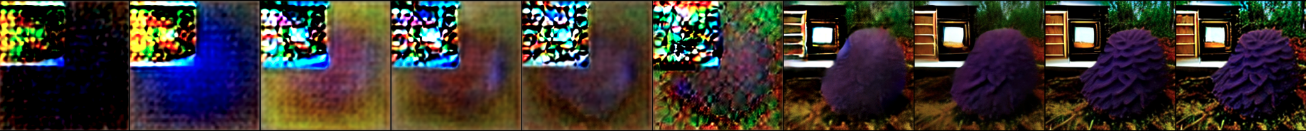} \\[5pt]
        \small \texttt{Model trained with CutMix --- seam artifact} \\[10pt]

        \includegraphics[width=0.6\linewidth]{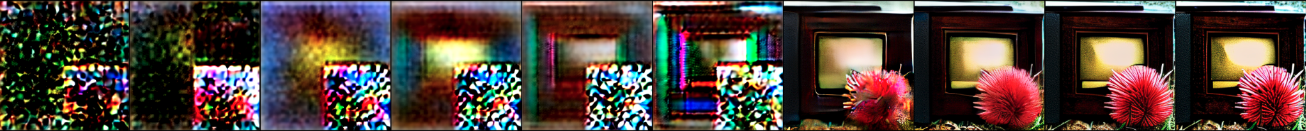} \\[5pt]
        \small \texttt{Model trained with CutMix --- blended scene adaptation} \\[10pt]
    \end{tabular}

    \caption{\textbf{CutMix failure case and learned adaptation.} Row~1: baseline without CutMix. Row~2: CutMix model producing a seam artifact. Row~3: CutMix model showing compositional scene adaptation. The $512^2$ fine-tuned model substantially reduces row-2-style artifacts.}
    \label{fig:cutmix_failurecase}
\end{figure*}

Figure~\ref{fig:cutmix_failurecase} exposes a failure mode of models trained with CutMix augmentation when confronted with prompts juxtaposing semantically disjoint concepts. Rather than synthesizing a coherent scene, these models sometimes decode the input as two independent images, resulting in visible artifacts: jagged seams, color bleeding, or distorted transitions along the composition boundary (Figure~\ref{fig:cutmix_failurecase}, row~2).

Intriguingly, CutMix-trained models also exhibit a learned adaptability to disjoint concepts. As seen in Figure~\ref{fig:cutmix_failurecase} (row~3), the model attempts to ``blend'' the scene to accommodate both prompts---e.g., combining indoor lighting with an outdoor object---rather than producing a hard boundary. This suggests the model has learned a form of scene-level compositionality even when it cannot fully reconcile the concepts.

Our fine-tuned $512^2$ model substantially reduces these discontinuities, generating more globally coherent images. This improvement suggests that higher-resolution training helps the model reconcile disjoint regions by leveraging finer spatial details.

\section{Additional Quantitative Results}
\label{app:quanti}

Table~\ref{tab:fid_TA_TApIA} reports a full image quality ablation study of the baseline, TA and TA+IA models on the in-domain \colorbox{YellowOrange!20}{ImageNet} validation set and the out-domain \colorbox{Orchid!20}{COCO} validation set.



Table~\ref{vic_tab:app_sota_aes} compares our best model against state-of-the-art models on aesthetic and human-preference metrics. Despite using significantly fewer parameters and orders-of-magnitude less training data, our model achieves competitive performances.

Tables~\ref{tab:miro_abla_bins} reports ablation of the number of reward bins used in the MIRO conditioning settin~\citep{dufour2025miro}, on the GenEval overall score and aesthetic metrics, complementing the comparison of Section~\ref{subsec:miro}.

\begin{table*}[!h]
\centering
\resizebox{\linewidth}{!}{
\begin{tabular}{
    l c c !{\color{gray!60}\vrule}
                c@{}c
                c@{}c
                c@{}c
                c@{}c
                c@{}c
                !{\color{gray!60}\vrule}
                c@{}c}
\toprule

\multirow{2}{*}{\textbf{Model}} &
\multirow{2}{*}{\textbf{TA}} &
\multirow{2}{*}{\textbf{IA}} &
\multicolumn{2}{c}{\textbf{FID Inc.}$\downarrow$} &
\multicolumn{2}{c}{\textbf{Prec.}$\uparrow$} &
\multicolumn{2}{c}{\textbf{Rec.}$\uparrow$} &
\multicolumn{2}{c}{\textbf{Den.}$\uparrow$} &
\multicolumn{2}{c!{\color{gray!60}\vrule}}{\textbf{Cov.}$\uparrow$} &
\multicolumn{2}{c}{\textbf{Jina CLIP}$\uparrow$} \\

& & &
\colorbox{YellowOrange!20}{IN-Val} & \colorbox{Orchid!20}{COCO} &
\colorbox{YellowOrange!20}{IN-Val} & \colorbox{Orchid!20}{COCO} &
\colorbox{YellowOrange!20}{IN-Val} & \colorbox{Orchid!20}{COCO} &
\colorbox{YellowOrange!20}{IN-Val} & \colorbox{Orchid!20}{COCO} &
\colorbox{YellowOrange!20}{IN-Val} & \colorbox{Orchid!20}{COCO} &
\colorbox{YellowOrange!20}{IN-Val} & \colorbox{Orchid!20}{COCO} \\
\midrule

\multirow{4}{*}{DiT-I}
& \textcolor{Red}{\faTimes} & \textcolor{Red}{\faTimes}
& 20.14 & 71.00
& 0.67 & 0.46
& 0.29 & 0.07
& 0.71 & 0.37
& 0.39 & 0.08
& 31.21 & 22.42 \\

& \textcolor{Green}{\faCheck} & \textcolor{Red}{\faTimes}
& 6.29 & 45.71
& 0.77 & 0.64
& 0.76 & 0.44
& 0.82 & 0.52
& 0.72 & 0.29
& 38.45 & 38.39 \\

& \textcolor{Green}{\faCheck} & \colorbox{Lavender!20}{Crop}
& 6.20 & 44.04
& 0.77 & 0.63
& 0.75 & 0.41
& 0.83 & 0.51
& 0.74 & 0.29
& 38.45 & 38.39 \\

& \textcolor{Green}{\faCheck} & \colorbox{Cerulean!20}{CutMix}
& 7.30 & 49.12
& 0.79 & 0.65
& 0.74 & 0.45
& 0.88 & 0.54
& 0.75 & 0.30
& 38.77 & 36.80 \\

\greyrule

\multirow{3}{*}{TextRIN-I}
& \textcolor{Red}{\faTimes} & \textcolor{Red}{\faTimes}
& 84.77 & 46.35
& 0.75 & 0.52
& 0.05 & 0.18
& 1.40 & 0.45
& 0.10 & 0.15
& 20.55 & 14.06 \\

& \textcolor{Green}{\faCheck} & \textcolor{Red}{\faTimes}
& 6.16 & 49.89
& 0.80 & 0.66
& 0.72 & 0.42
& 0.89 & 0.61
& 0.76 & 0.28
& 38.01 & 37.85 \\

& \textcolor{Green}{\faCheck} & \colorbox{Cerulean!20}{CutMix}
& 6.62 & 49.31
& 0.80 & 0.66
& 0.70 & 0.41
& 0.90 & 0.57
& 0.76 & 0.29
& 38.17 & 37.71 \\

\bottomrule
\end{tabular}}

\caption{\textbf{Image quality} of AIO, TA, and TA {+} IA models w.r.t. in-domain Imagenet validation set (\colorbox{YellowOrange!20}{IN-Val}) and out-domain MSCOCO-30K validation set (\colorbox{Orchid!20}{COCO}). FID reported is FID \texttt{Inception v3}. Precision, Recall, Density and Coverage are computed using \texttt{DINOv2} features.}
\label{tab:fid_TA_TApIA}
\end{table*}

\begin{table*}[t]
    \centering
        \begin{tabular}{l@{\;}c@{\;}c@{\;}!{\color{gray!60}\vrule}c@{\;}c@{\;}c@{\;}c@{\;}}
            \toprule
             \textbf{Model}  & \textbf{\#params} & \makecell[c]{\textbf{\#train} \\ \textbf{data}} & \textbf{Aes. Score$\uparrow$} & \textbf{PickScore$\uparrow$} & \textbf{HPSv2.1$\uparrow$} & \textbf{ImageReward$\uparrow$}\\
            \midrule
             SD v1.5 & 0.9B & 5B+ & 5.68 & 21.3 & 0.25 & 0.24 \\
             SD v2.1 & 0.9B & 5B+ & 5.81 & 21.5 & 0.26 & 0.38 \\
             PixArt-$\alpha$ & 0.6B & 25M & \underline{6.47} & 22.6 & \underline{0.29} & 0.97 \\
             PixArt-$\Sigma$ & 0.6B & 35M+ & 6.44 & 22.5 & \underline{0.29} & 1.02 \\
             CAD & 0.35B & - & 5.56 & 21.4 & 0.26 & 0.69 \\
             Sana-0.6B & 0.6B & - & 6.31 & \underline{22.8} & \textbf{0.30} & \textbf{1.23} \\
             Sana-1.6B & 1.6B & - & 6.36 & \underline{22.8} & \textbf{0.30} & \textbf{1.23} \\
             SDXL & 2.6B & 5B+ & 5.94 & 22.0 & 0.25 & 0.46 \\
             SD3-Medium & 2B & 1B+ & 6.18 & 22.5 & 0.30 & 1.15 \\
             FLUX-dev & 12B & - & \textbf{6.56} & \textbf{22.9} & \textbf{0.30} & \underline{1.19} \\
             \greyrule
             \rowcolor{Periwinkle!10} SiT-I H CutMix MIRO & 1B & 1.2M & 5.77 & 20.84 & 0.28 & 0.73  \\
            \bottomrule
        \end{tabular}
    \caption{\textbf{Results on Reward Metrics.} Results are computed using the \texttt{PartiPrompts}~\cite{yu2022scalingautoregressivemodelscontentrich}. SOTA scores are computing using HuggingFace checkpoints at their native resolution. \textbf{Bold} indicates best, \underline{underline} second best.}
    \label{vic_tab:app_sota_aes}
\end{table*}

\begin{table}[h]
\centering
\resizebox{\linewidth}{!}{
    \begin{tabular}{l c c c c c}
        \toprule
        \textbf{Num.\ bins} & \textbf{GenEval (Overall)$\uparrow$} & \textbf{PickScore$\uparrow$} & \textbf{Aes.Score$\uparrow$} & \textbf{HPSv2.1$\uparrow$} & \textbf{ImageReward$\uparrow$} \\
        \midrule
        16 & 0.51 & 19.77 & 5.22 & 0.22 & 0.04 \\
        64 & 0.47 & 19.67 & 5.13 & 0.21 & -0.03 \\
        \bottomrule
    \end{tabular}}
\caption{\textbf{Ablation on the number of reward bins} used for MIRO conditioning, for SiT-I at 225k steps.}
\label{tab:miro_abla_bins}
\end{table}


\section{Full Implementation Details}
\label{app:train_arch_details}

Figure~\ref{fig:architectures} shows the architecture schematic for DiT-I and TextRIN-I, introduced in Section~\ref{sec:setup}; SiT-I reuses the DiT-I schematic (see below). Below we give key numerical hyperparameters for all four architectures and the full textual architecture description.

\begin{figure}[h!]
    \centering
    \begin{subfigure}[c]{0.28\textwidth}
        \centering
        \resizebox{\linewidth}{!}{\input{images/dit_architecture}}
    \end{subfigure}
    \hfill
    \vrule
    \hfill
    \begin{subfigure}[c]{0.6\textwidth}
        \centering
        \resizebox{\linewidth}{!}{\input{images/rin_architecture}}
    \end{subfigure}
    \caption{\textbf{Architecture blocks used in our experiments.} \textit{Left}: DiT-I block, also used by SiT-I. \textit{Right}: TextRIN-I block.}
    \label{fig:architectures}
\end{figure}

\subsection*{Key architecture hyperparameters}
\label{app:arch_params}

\begin{figure}[ht]
    \centering
    \begin{minipage}{0.28\linewidth}
        \centering
        \begin{tabular}{lc}
            \toprule
             \textbf{Parameter} & \textbf{Value} \\
            \midrule
             Num.\ blocks                   & 24 \\
             Embedding dim.                 & 1024 \\
             Num.\ heads                    & 16 \\
             Patch size                     & 2 \\
             Input size (256$^2$)           & $(4,32,32)$ \\
             Text registers                 & 16 \\
             Cond.\ self-attn blocks        & 2 \\
             Cond.\ self-attn heads         & 16 \\
             Cond.\ FFN expansion           & 4 \\
             DiT FFN expansion              & 4 \\
             Uncond.\ drop prob.            & 0.1 \\
            \bottomrule
        \end{tabular}
        \caption*{\textbf{DiT-I / SiT-I parameters}}
    \end{minipage}
    \hfill
    \begin{minipage}{0.28\linewidth}
        \centering
        \begin{tabular}{lc}
            \toprule
             \textbf{Parameter} & \textbf{Value} \\
            \midrule
             Input size (256$^2$)           & $(4,32,32)$ \\
             Num.\ latents                  & 128 \\
             Latent dim.                    & 512 \\
             Num.\ proc.\ layers            & 2 \\
             Num.\ blocks                   & 3 \\
             Patch size                     & 2 \\
             Read/Write heads               & 8 \\
             Compute heads                  & 16 \\
             Data pos.\ emb.                & learned \\
             Text registers                 & 16 \\
             Uncond.\ drop prob.            & 0.1 \\
            \bottomrule
        \end{tabular}
        \caption*{\textbf{TextRIN-I parameters}}
    \end{minipage}
    \hfill
    \begin{minipage}{0.28\linewidth}
        \centering
        \begin{tabular}{lc}
            \toprule
             \textbf{Parameter} & \textbf{Value} \\
            \midrule
             Num.\ blocks                   & 24 \\
             Embedding dim.                 & 1024 \\
             Num.\ heads                    & 16 \\
             Patch size                     & 16 \\
             Input size (256$^2$)           & $(3,256,256)$ \\
             Bottleneck dim.                & 128 \\
             Rotary position emb.           & \faCheck \\
             Text registers                 & 16 \\
             Cond.\ self-attn blocks        & 4 \\
             FFN activation                 & SwiGLU \\
             Uncond.\ drop prob.            & 0.1 \\
            \bottomrule
        \end{tabular}
        \caption*{\textbf{JiT-I parameters}}
    \end{minipage}
    \caption{\textbf{Key architecture parameters} of DiT-I/SiT-I (left), TextRIN-I (middle), and JiT-I (right). DiT-I and SiT-I share the same architecture and hyperparameters, differing only in their training objective (see main text).}
    \label{fig:arch_params}
\end{figure}

\vspace{2mm}
\noindent \textbf{DiT-I.} We use DiT~\citep{peebles2023scalable} adapted for text-conditional generation by replacing \verb|AdaLN-Zero| conditioning with \textit{cross-attention}, as in PixArt-$\alpha$~\citep{chen2023pixart}. Before cross-attention, the text condition is refined using \textit{two} self-attention layers. Following~\citet{esser2024scaling}, we add QK-Normalization~\citep{henry2020query} to all self-attention and cross-attention blocks, and LayerScale~\citep{touvron2021going} to all residual blocks for training stability.

\vspace{2mm}
\noindent \textbf{TextRIN-I.} We use the off-the-shelf CAD architecture from~\citet{dufour2024don}, an adaptation of RIN~\citep{jabri2023scalableadaptivecomputationiterative} for text-conditional generation. Full architectural details are provided in the appendix of~\citet{dufour2024don}.

\vspace{2mm}
\noindent \textbf{SiT-I.} SiT-I reuses the exact DiT-I backbone described above (same block count, embedding dimension, cross-attention text-conditioning, QK-Normalization and LayerScale; see the DiT-I column of Figure~\ref{fig:arch_params}). The only difference is the training objective: SiT-I is trained with the flow-matching objective of~\citet{ma2024eccvSIT} in place of the diffusion objective (Equation~\ref{eq:diff}), using log-normal timestep sampling and an Euler ODE sampler at inference.

\vspace{2mm}
\noindent \textbf{JiT-I.} JiT-I adapts JiT~\citep{li2025jit} for text-conditional generation, following the text-conditioning recipe of PixelGen~\citep{ma2026pixelgen}. Unlike DiT-I, TextRIN-I, and SiT-I, JiT-I operates directly on raw pixels: patches are linearly embedded through a low-rank bottleneck projection instead of a VAE latent, image tokens use rotary position embeddings, and text tokens are refined by four time-conditioned self-attention blocks before being jointly attended to by the image tokens within the same attention operation (as opposed to a separate cross-attention block). JiT-I is trained with a flow-matching objective and sampled with a Heun ODE solver. Key hyperparameters are given in Figure~\ref{fig:arch_params}.

\vspace{2mm}
\noindent \textbf{Shared components.} DiT-I, TextRIN-I, and SiT-I all operate in the latent space of the latent diffusion framework~\citep{rombach2022high}: images are encoded with the pre-trained VAE from Stable Diffusion (\href{https://huggingface.co/stabilityai/sd-vae-ft-ema}{\texttt{stabilityai/sd-vae-ft-ema}}). JiT-I instead operates directly in pixel space and uses no VAE. Text is encoded with T5-XL (\href{https://huggingface.co/google/flan-t5-xl}{\texttt{google/flan-t5-xl}}) for all four architectures.


\section{Full Captioning Details}
\label{app:captions}

Key aspects of our captioning strategy (the LLaVA token-pruning trick and the CutMix-specific prompts) are summarized in Section~\ref{sec:method}. Full details are provided below.

\paragraph{Captioning efficiently with LLaVA} To caption images, we use the checkpoint \texttt{llama3-llava-next-8b-hf} (available on HuggingFace: \url{https://huggingface.co/llava-hf/llama3-llava-next-8b-hf})  with the prompt \textit{"Describe this image"}. 
LLaVA encodes images using a dynamic resolution scheme. It processes both the entire image and four distinct patches as unique images and concatenates them. For 256x256 images, LLaVA uses around 2500 image tokens. To make the captioning process more efficient, we prune the image tokens, retaining only the tokens of the entire image and discarding patch-specific tokens. This optimization increased inference speed by a factor of 2.7, without compromising performances. Examples of long captions generated by LLaVA are given in Figure \ref{fig:fig_captions_abla}.

\paragraph{Captioning CutMix images} We caption CutMix images from CM$^{1/2}$ with similar settings used for captioning the original ImageNet images. However, to ensure that LLaVA does not describe both the base and the CutMix images independently, we use a different prompt: \textit{“Describe this image. Consider all the objects in the picture. Describe them, describe their position and their relation. Do not consider the image as a composite of images. The image is a single scene image”}.

For settings CM$^{1/4}$, CM$^{1/9}$ and CM$^{1/16}$, LLaVA tends to either ignore the smaller CutMix image or describe the image as a composite of two images. To avoid this behaviour, we encode the image by using the entire image patch and add tokens from the patch to which the CutMix image belongs. We use the following prompt: \textit{“Describe this image. Consider all the objects in the picture. Describe them, describe their position and their relation. Do not consider the image as a composite of images. The image is a single scene image”}. Examples of long captions generated by LLaVA for CutMix images are given in Figure \ref{fig:fig_captions_abla}.


\section{Full CutMix Details}
\label{app:cutmix}

A summary of the CutMix hyperparameter ablations is presented in Section~\ref{sec:method}. This appendix provides the modified training losses, the threshold-gated Bernoulli sampling scheme. 

The CutMix framework systematically combines concepts while preserving object centrality. Our framework defines four precise augmentation patterns, each designed to maintain visual coherence while introducing novel concept combinations. These are briefly described below: 

\begin{enumerate}  
\item \textbf{CM$^{1/2}$} (Half-Mix):\\
    \textit{Scale:} Both images maintain their original resolution.\\
    \textit{Position:} Deterministic split along height or width.\\
    \textit{Coverage:} Each concept occupies 50\% of final image.\\
    \textit{Preservation:} Both concepts maintain full resolution.
\item \textbf{CM$^{1/4}$} (Quarter-Mix):\\
    \textit{Scale:} CutMix image resized to 50\% side length.\\
    \textit{Position:} Fixed placement at one of four corners.\\
    \textit{Coverage:} 2nd concept occupies 25\% of final image.\\
    \textit{Preservation:} Base image center region remains intact.

\item \textbf{CM$^{1/9}$} (Ninth-Mix):\\
    \textit{Scale:} CutMix image resized to 33.3\% side length.\\
    \textit{Position:} Fixed placement along image borders.\\
    \textit{Coverage:} 2nd concept occupies 11.1\% of final image.\\
    \textit{Preservation:} Base image center, corners remain intact.

\item \textbf{CM$^{1/16}$} (Sixteenth-Mix):\\
    \textit{Scale:} CutMix image resized to 25\% side length.\\
    \textit{Position:} Random placement not central 10\% region.\\
    \textit{Coverage:} 2nd concept occupies 6.25\% of final image.\\
    \textit{Preservation:} Base image center region remains intact.
\end{enumerate}

\begin{figure*}[t]
    \centering
    \resizebox{\textwidth}{!}{\begingroup

\renewcommand{\ttdefault}{pcr}
\centering
\scriptsize

\begin{tabular}{m{4cm}m{12cm}}
    \toprule
    \multicolumn{1}{c}{\textbf{Setting}} & \multicolumn{1}{c}{\textbf{Long Caption}} \\
    \midrule

    \begin{center}
        \textbf{LLaVA w/o CM}
        \begin{tikzpicture}
        \node at (0,1.5,0) {\includegraphics[height=2.95cm, width=2.95cm]{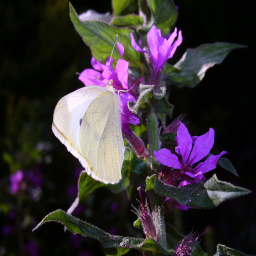}};
        \end{tikzpicture}
    \end{center}
    & 
    \texttt{
        In the heart of the verdant background, a delicate white butterfly with a hint of yellow around its edges is captured in mid-flight. Its wings, a blend of two shades of cream and white, add a soft contrast to the vibrant colors of the flower it's hovering over. The flower, a stunning shade of purple, is adorned with a yellow center that speaks of the sun's rays it might have been basking in. Surrounding it is a lush green foliage, adding depth to this tranquil scene. The butterfly, positioned slightly to the left of the flower's center, seems to be in the process of alighting or perhaps taking off, adding a dynamic element to this otherwise serene tableau.
    } \\
    
    \midrule
    \begin{center}
        \textbf{CM$^{1/2}$}
        
        \begin{tikzpicture}
        \node at (0,1.5,0) {\includegraphics[height=2.95cm, width=2.95cm]{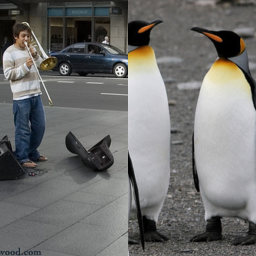}};
        \end{tikzpicture}
    \end{center}
     & 
    \texttt{
        On the left side, there is a person playing the trumpet on a street. The individual is standing on the ground, wearing a long-sleeved shirt and jeans. On the ground next to the player, there is a trash can. The street is made of asphalt, with buildings in the background and a clear sky overhead. On the right side of the image, there are two penguins standing on a gravel surface, possibly a road or a rocky beach. The penguins are frozen in place, displaying a contrast to the dynamic scene on the left side. The penguins' yellow beaks, black bodies, and white feathers are clearly visible. The image captures an everyday moment of music played on the street, juxtaposed with the surreal and unexpected presence of the penguins. The image does not provide information about the time of day, specific location, or any actions being performed by the penguins or the person. The image is static, presenting two moments captured in time rather than a sequence of events.
    } \\
    
    \midrule
    \begin{center}
        \textbf{CM$^{1/4}$}
        
        \begin{tikzpicture}
        \node at (0,1.5,0) {\includegraphics[height=2.95cm, width=2.95cm]{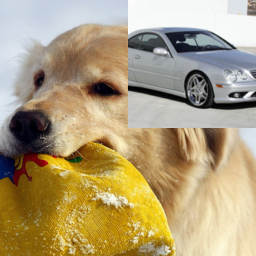}};
        \end{tikzpicture}
    \end{center}
     & 
    \texttt{
        The image shows a large, muscular dog in the foreground and a silver sports car in the background. The dog, appearing to be a Golden Retriever, is on the left side of the frame, occupying a significant portion of the space. It has a yellow hue with darker markings, and its ears are floppy. The dog is heavily coated with snowflakes, suggesting that the photo was taken in a snowy environment. The sports car, positioned on the right, is a two-door coupe with sleek curves and a notable design, featuring the Mercedes-Benz logo on its front grille. The car has a silver finish, and the photo captures it from a perspective that shows the front and side profile. The car is parked on an asphalt surface, possibly a parking lot or a driveway. The dog is facing the camera with direct gaze, while the car is positioned slightly towards the side, away from the viewer's perspective.
    } \\
    
    \midrule
    \begin{center}
        \textbf{CM$^{1/9}$}
        
        \begin{tikzpicture}
        \node at (0,1.5,0) {\includegraphics[height=2.95cm, width=2.95cm]{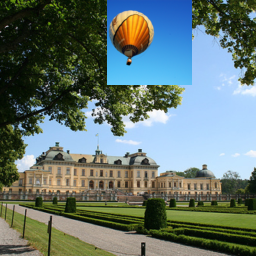}};
        \end{tikzpicture}
    \end{center}
     & 
    \texttt{
        The image depicts a picturesque outdoor scene featuring an ornate building, which appears to be a palace or manor house, with classical architectural elements including symmetrical windows, a central cupola, and multiple chimneys. In front of the building is a well-maintained garden with pathways and neatly trimmed hedges or borders. Above the garden, there is a clear blue sky with a few scattered clouds. In the sky, there is a single hot air balloon with a bright orange and yellow pattern. The balloon is floating at a considerable height above the garden and the building, suggesting it might be part of a leisure activity or a special event. The image is a photograph with natural lighting, indicative of a sunny day.
    } \\
    
    \midrule
    \begin{center}
        \textbf{CM$^{1/16}$}
        
        \begin{tikzpicture}
        \node at (0,1.5,0) {\includegraphics[height=2.95cm, width=2.95cm]{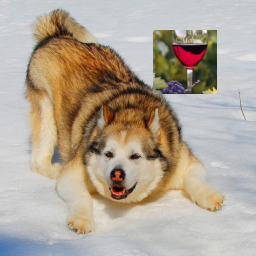}};
        \end{tikzpicture}
    \end{center}
     & 
    \texttt{
        The image is a photograph featuring a husky dog resting in the snow. The dog has a light coat with darker markings around its face and ears, and it is lying on its side with its head up, looking directly at the camera. Its eyes are open and its mouth is slightly open, showing teeth and a pink tongue, which suggests the dog might be panting or in a relaxed state. Next to the dog's side, there is a wine glass with red wine and a few purple flowers, which could be lilacs, positioned on the left side of the glass stem. The wine glass and flowers are set against a blurred background that gives the impression of greenery.
    } \\
    
    \bottomrule
\end{tabular}


\endgroup}
    \caption{\textbf{Long captions generated by our synthetic LLaVA captioner}. The captions generated are highly diverse and add in much more intricate details of \textit{compositionality}, \textit{colors} as well as \textit{concepts} which are not present in the original ImageNet dataset. The captions generated for our augmented images are also highly coherent and explain the scene in a much more realistic way.}
    \label{fig:fig_captions_abla}
\end{figure*}

Each augmentation strategy generates 1,281,167 samples, matching ImageNet's training set size. Figure \ref{fig:fig_captions_abla} shows examples of the different structured augmentations.

We also define \textbf{CM$^{all}$}, which uniformly samples from all four patterns. The CM$^{all}$ variant combines equal proportions (25\%) from each pattern to maintain the same total sample count.
Post-augmentation, we apply LLaVA captioning to all generated images, ensuring semantic alignment between visual and textual representations. This produces detailed descriptions that accurately reflect the augmented content while maintaining natural language fluency.

\subsection{CutMix Training Pipeline}
\label{subapp:cutmix_training}

Because CutMix images have sharp composition boundaries, training on them at all noise levels would cause the model to reproduce the boundary artifacts. We prevent this by using CutMix images only once the corrupted input is noisy enough that the artifact is no longer visible. Concretely, let $\mathcal{A}$ be the original dataset and $\mathcal{A}_\text{IA}$ the CutMix-augmented dataset. We introduce a random variable $\rho$ sampled conditionally on the timestep $t$ via a threshold-gated Bernoulli:
\begin{align}
 \mathcal{B}_{\tau,p}(t) = \begin{cases}
        0, & t \leq \tau,\\
        \mathcal{B}_p, & \text{else,}
    \end{cases}
\end{align}
where $\mathcal{B}_p$ is a Bernoulli distribution with parameter $p$. The text-image pair $(x_0, c)$ is then sampled from the dataset conditioned on $\rho$:
\begin{align}
    \mathcal{A}(\rho) = \begin{cases}
        \mathcal{A}, & \rho = 0,\\
        \mathcal{A}_\text{IA}, & \rho = 1.
    \end{cases}
    \label{eq:augdiff_dataset}
\end{align}
This sampling scheme is shared by our diffusion (DiT-I, TextRIN-I) and flow-matching (SiT-I) architectures; only the noising process and the training objective it feeds into differ.

\textit{Diffusion.} For DiT-I and TextRIN-I, we replace the standard diffusion loss (Equation~\ref{eq:diff}) with:
\begin{align}
    \min_\theta \mathbb{E}_{\substack{t\sim\mathcal{U}(0,T),\\ \rho\sim\mathcal{B}_{(\tau,p)}(t),\\ (x,c)\sim\mathcal{A}(\rho),\\ \epsilon\sim\mathcal{N}(0,1)}}\left[ \|\epsilon - \epsilon_\theta(x_t, c, t)\|^2  \right],
    \label{eq:augdiff}
\end{align}
where the noisy image $x_t {=} \sqrt{\gamma(t)} x_0 + \sqrt{1-\gamma(t)} \epsilon$ is as usual.

\textit{Flow matching.} For SiT-I, we apply the same conditional sampling scheme to the flow-matching loss (Equation~\ref{eq:fm}) instead:
\begin{align}
    \min_\theta \mathbb{E}_{\substack{t\sim\mathcal{U}(0,1),\\ \rho\sim\mathcal{B}_{(\tau,p)}(t),\\ (x,c)\sim\mathcal{A}(\rho),\\ \epsilon\sim\mathcal{N}(0,1)}}\left[ \|(\epsilon - x_0) - v_\theta(x_t, c, t)\|^2  \right],
    \label{eq:augfm}
\end{align}
where the interpolated sample $x_t {=} (1-t) x_0 + t \epsilon$ follows the linear path of Equation~\ref{eq:fm}, and the threshold $\tau$ is expressed on the flow-matching timestep scale $[0,1]$ rather than the diffusion scale $[0,T]$.

In both cases, the timestep $t$ is sampled uniformly, and the loss is easy to implement as it amounts to a conditional data sampling scheme during mini-batch construction. Figure~\ref{fig:cutmix_pipeline} illustrates the complete CutMix data curation and training pipeline.

\begin{figure*}[t]
    \centering
    \resizebox{\linewidth}{!}{\input{images/figure_2_pipeline}}
    \caption{\textbf{Pipeline of our CutMix data curation and training process.} Starting from ImageNet, we (a) use LLaVA to caption original images into long detailed captions (top branch), and (b) use CutMix strategies to create new composite images and caption them with LLaVA using scene-aware prompts (bottom branch). During training, batches mix original and CutMix images, gated by timestep $t$ and probability $p$.}
    \label{fig:cutmix_pipeline}
\end{figure*}

\subsection{Full CutMix Augmentation Ablations}

We report ablations on the three key CutMix hyperparameters: the CutMix setting, the probability $p$ of sampling a CutMix image, and the threshold $\tau$. Based on these ablations, we select \CALL, $p{=}0.25$ and $\tau{=}400$ for all final models.

\subsubsection{Ablation on CutMix settings}
\label{subsec:abla_setting}

First, we analyse the performances of the pixel augmentations for $\{$\CONE, \CTWO, \CTHREE, \CFOUR, \CALL $\}$ settings. We fix the probability of using a pixel-augmented image in the batch when $t>\tau$ to $p=0.5$ and we measure both image quality and composition ability. Results are reported in Table \ref{tab:abl_cutmix_setting}.

For image quality, all settings seem to perform similarly, with \CONE being the best at 6.13 FID and \CALL being the worst at 6.81 FID. This indicates that all settings are able to avoid producing uncanny images that would disturb the training too much.

For composition ability, \CFOUR can improve over the baseline on extended prompts, whereas \CALL can improve over the baseline on original prompts. Overall, only \CALL manages to keep closer performances between the original prompts and the extended ones. Since \CALL is a mixture of all other settings, it also has the most diverse training set and is thus harder to overfit. As such, we consider \CALL for the best models.


\begin{table}[ht]
    \centering
    \small
    \begin{tabular}{l@{\;}c@{\;}!{\color{gray!60}\vrule}c@{\;}!{\color{gray!60}\vrule}c@{\;}c@{\;}}
        \toprule
        \multirow{2}{*}{\textbf{Model}} & \multirow{2}{*}{\makecell{\textbf{CutMix} \\ \textbf{Settings}}} & \multirow{2}{*}{\textbf{FID$\downarrow$}} & \multicolumn{2}{c}{\textbf{GenEval$\uparrow$}} \\
        & & & $\diamond$ & $\star$ \\
        \midrule
        \multirow{5}{*}{\rotatebox{90}{CAD-I}} 
          & CM$^{1/2}$ & 6.13 & 0.46 & 0.55 \\
          & CM$^{1/4}$ & 6.41 & 0.49 & 0.53 \\
          & CM$^{1/9}$ & 6.63 & 0.51 & 0.51 \\
          & CM$^{1/16}$ & 6.42 & 0.47 & 0.56 \\
          & CM$^{all}$ & 6.81 & 0.53 & 0.55 \\
        \bottomrule
    \end{tabular}
    \caption{\textbf{Ablation study} on CutMix settings. The probability of sampling CutMix images used here is $\rho=0.5$. Models are trained for 250k steps. FID is computed on the ImageNet val set with long prompts, using the \texttt{Inception-v3} backbone. $\diamond$ means original GenEval prompts. $\star$ means extended GenEval prompts.}
    \label{tab:abl_cutmix_setting}
\end{table}


\begin{table}[ht]
    \centering
    \small
    \begin{tabular}{l@{\;}c@{\;}!{\color{gray!60}\vrule}c@{\;}!{\color{gray!60}\vrule}c@{\;}c@{\;}}
        \toprule
        \multirow{2}{*}{\textbf{Model}} & \multirow{2}{*}{$\rho$} & \multirow{2}{*}{\textbf{FID$\downarrow$}} & \multicolumn{2}{c}{\textbf{GenEval$\uparrow$}} \\
        & & & $\diamond$ & $\star$ \\
        \midrule
        \multirow{5}{*}{\rotatebox{90}{CAD-I}} 
          & 0 & 6.16 & 0.51 & 0.55 \\
          & 0.25 & 5.99 & 0.55 & 0.58 \\
          & 0.5 & 6.41 & 0.49 & 0.53 \\
          & 0.75 & 6.71 & 0.45 & 0.53 \\
          & 1 & 6.07 & 0.48 & 0.49 \\
        \bottomrule
    \end{tabular}
    \caption{\textbf{Ablation study} on probability $\rho$ of sampling a CutMix image during training. The CutMix setting is CM$^{1/4}$. Models are trained for 250k steps. FID is computed on ImageNet val set with long prompts, using the \texttt{Inception-v3} backbone. $\diamond$ means original GenEval prompts. $\star$ means extended GenEval prompts.}
    \label{tab:abl_cutmix_proba}
\end{table}

\begin{table}[ht]
    \centering
    \small
    \begin{tabular}{l@{\;}c@{\;}!{\color{gray!60}\vrule}c@{\;}!{\color{gray!60}\vrule}c@{\;}c@{\;}}
        \toprule
        \multirow{2}{*}{\textbf{Model}} & \multirow{2}{*}{$\tau$} & \multirow{2}{*}{\textbf{FID$\downarrow$}} & \multicolumn{2}{c}{\textbf{GenEval$\uparrow$}} \\
        & & & $\diamond$ & $\star$ \\
        \midrule
        \multirow{4}{*}{\rotatebox{90}{CAD-I}} 
          & 300 & 6.99 & 0.51 & 0.53 \\
          & 400 & 6.62 & 0.55 & 0.57 \\
          & 500 & 6.16 & 0.48 & 0.55 \\
          & 600 & 5.90 & 0.50 & 0.55 \\
        \bottomrule
    \end{tabular}
    \caption{\textbf{Ablation study} on timestep threshold $\tau$. The CutMix setting is \CALL. Models are trained for 250k steps. FID is computed on ImageNet val set with long prompts, using the \texttt{Inception-v3} backbone. $\diamond$ means original GenEval prompts. $\star$ means extended GenEval prompts.}
    \label{tab:abla_tau_cutmix}
\end{table}

\subsubsection{Ablation on CutMix probability}
\label{subsec:abla_proba}

Next, we analyse the influence of the probability $p$ of using a pixel augmented image in the batch, when the condition on $t$ is met.
Results for $p\in \{0.25, 0.5, 0.75, 1.0\}$ are shown in Table \ref{tab:abl_cutmix_proba}, using \CTWO pixel augmentations.

As we can see in terms of image quality, the FID is slightly degraded by having too frequent pixel augmentation ($p> 0.5$). This can be explained by the fact that pixel-augmented images are only seen when $t>\tau$. As such, a high value for $p$ creates a distribution gap between the images seen for $t>\tau$ and the images seen for $t\leq \tau$.

Composition ability shows a similar behaviour with the GenEval overall score decreasing when $p$ increases for both the original and the extended prompts. As such, we select $p{=}0.25$ for the best trade-off between diversity and image quality.

\subsubsection{Ablation on threshold $\tau$}
\label{subsec:abla_tau}

Finally, we analyse the influence of the threshold $\tau$, which enables CutMix images to be sampled in training batches. Table \ref{tab:abla_tau_cutmix} shows the FID Inception on ImageNet Val and the GenEval scores of models trained with different $\tau$ values.

We find that $\tau=400$ results in the highest GenEval score of $0.55$ on original prompts and $0.57$ on extended prompts, while $\tau=600$ yields the lowest FID on ImageNet Val. As such, we use $\tau=400$ for the best models.


\section{Cropping details}
\label{app:cropping}

Our cropping training methodology (see Figure \ref{fig:crop_pipeline}) removes spurious concept correlations due to its masking scheme. We maintain the original captions and force the model to independently identify relevant textual elements.  This creates a more challenging learning task for the model that enhances text-image alignment. During training, we only consider tokens corresponding to small portion of the image and mask out the rest from both the loss function and cross-attention layers. Given we do this online, this is highly efficient and allows an infinite training set based on ImageNet to.

For making the model understand the full dynamics of the training data, in our training scheme with crops, we only feed cropped images to the model with a probability, $p=0.5$. In rest of the cases, we use entire image. To keep the training scheme as simple as possible, for cropped versions, we only use crop resolution of >50\% of the normal resolution.

\begin{figure*}[h!]
    \centering
    \resizebox{\linewidth}{!}{\input{images/crop_pipeline}}
    \caption{\textbf{Pipeline of our Cropping Data Curation and Training process.} Starting from ImageNet, we a) use LLaVa VLM to caption the images into long detailed caption 
    and b) use cropping strategies to create new images from ImageNet by cropping. We keep the same captions as if we were using the original image. During training, we do cropping online with a probability $p$ of sampling cropped images. The crop images can have any resolution >50\% of the original resolution.}
    \label{fig:crop_pipeline}
\end{figure*}

\end{document}